\documentclass{article}

\usepackage{microtype}
\usepackage{graphicx}
\usepackage{subcaption}
\usepackage{booktabs} 
\usepackage{pifont}   
\usepackage{amssymb}  
\usepackage{bm}       
\usepackage{amsmath}  
\usepackage{xcolor}   
\usepackage{tabularray}
\newcommand{\cmark}{\textcolor{green!60!black}{\text{\ding{51}}}}

\newcommand{\xmark}{\textcolor{red}{\text{\ding{55}}}}

\newcommand{\tmark}{\textcolor{orange}{\ensuremath{\bm{\triangle}}}}

\usepackage{enumitem} 

\usepackage{hyperref}
\usepackage{empheq}

\usepackage{multirow}
\usepackage{tabularx}
\usepackage{colortbl}
\usepackage{xcolor}
\usepackage{stfloats}
\usepackage{makecell}
\usepackage{longtable}
\usepackage{booktabs}
\usepackage{tabularx}
\usepackage{xcolor}

\usepackage[preprint]{icml2026}

\usepackage{amsmath}
\usepackage{amssymb}
\usepackage{mathtools}
\usepackage{amsthm}

\usepackage[most]{tcolorbox} 

\usepackage[capitalize,noabbrev]{cleveref}

\theoremstyle{plain}

\theoremstyle{definition}

\theoremstyle{remark}

\usepackage[textsize=tiny]{todonotes}

\icmltitlerunning{EgoPro-Bench: Benchmarking Personalized Proactive Interaction in Egocentric Video Streams}

\begin{document}

\twocolumn[
  \icmltitle{EgoPro-Bench: Benchmarking Personalized Proactive Interaction in Egocentric Video Streams}



  \icmlsetsymbol{equal}{*}

  \begin{icmlauthorlist}
    \icmlauthor{Dongchuan Ran}{equal,sensetime}
    \icmlauthor{Linyu Ou}{equal,sensetime,beiligong}
    \icmlauthor{Xueheng Li}{equal,sensetime,zhongkeda}
    \icmlauthor{Wenwen Tong}{sensetime}
    \icmlauthor{Chenxu Guo}{sensetime}
    \icmlauthor{Hewei Guo}{sensetime}
    \icmlauthor{Kaibing Wang}{sensetime}
    \icmlauthor{Lewei Lu}{sensetime}
  \end{icmlauthorlist}

  \icmlaffiliation{sensetime}{Sensetime Reasearch}
  \icmlaffiliation{beiligong}{Beijing Institute of Technology}
  \icmlaffiliation{zhongkeda}{University of Science and Technology of China}

  \icmlcorrespondingauthor{Lewei Lu}{lulewei@sensetime.com}
  
  \icmlkeywords{Machine Learning, ICML}

  \vskip 0.3in
    {
    \centering
    \includegraphics[
        width=0.97\textwidth,
        height=0.36\textheight,
    ]{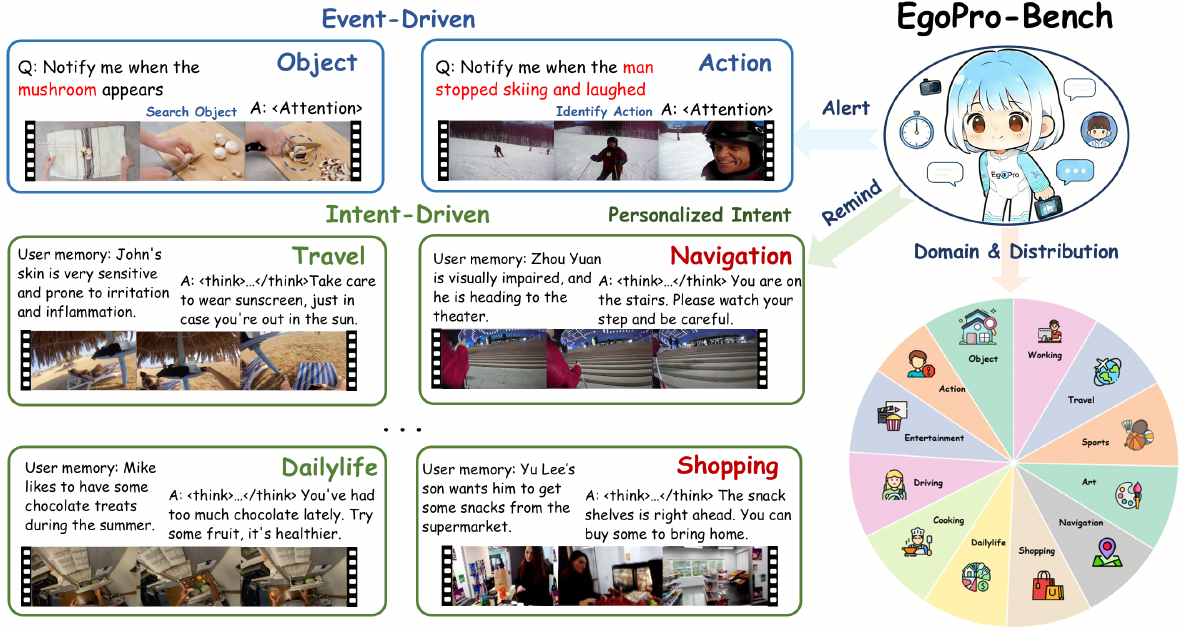} 
    \captionof{figure}{Examples and data distribution of EgoPro-Bench. The benchmark consists of two main categories (event-driven and intent-driven) and covers 12 distinct domains for personalized proactive interaction.}
    \label{fig:teaser}
  }
  \vskip 0.3in
]



\printAffiliationsAndNotice{}  

\begin{abstract}
Existing Multimodal Large Language Models (MLLMs) remain primarily reactive, failing to continuously perceive environments or proactively assist users. While emerging benchmarks address proactivity, they are largely confined to alert scenarios, neglect personalized context, and fail to evaluate the precise timing of human–machine interactions (HMI).
In this paper, we introduce EgoPro-Bench, a novel benchmark for training and evaluating proactive interaction capabilities based on streaming egocentric videos; it comprises 2,400 videos in the evaluation set and over 12,000 videos in the training set.
Unlike previous works, EgoPro-Bench leverages simulated user profiles to generate diverse user intentions and to construct high-fidelity HMI data across 12 distinct domains.
Subsequently, we propose a specialized evaluation protocol and metrics, train proactive interaction models designed for efficient reasoning and low-latency interaction on streaming video data, and conduct comprehensive evaluations.
Furthermore, we introduce an interaction principle termed ``short thinking, better interaction,'' which allocates a limited token budget prior to intent recognition, thereby enhancing interaction performance.
The experiments demonstrate that EgoPro-Bench substantially enhances the intention understanding capabilities of MLLMs and enables accurate identification of appropriate timings for HMI, thereby laying a solid foundation for next-generation user-centric proactive interactive agents.
\end{abstract}

\section{Introduction}
\begin{figure}[ht]
  \centering
  \includegraphics[width=\columnwidth]{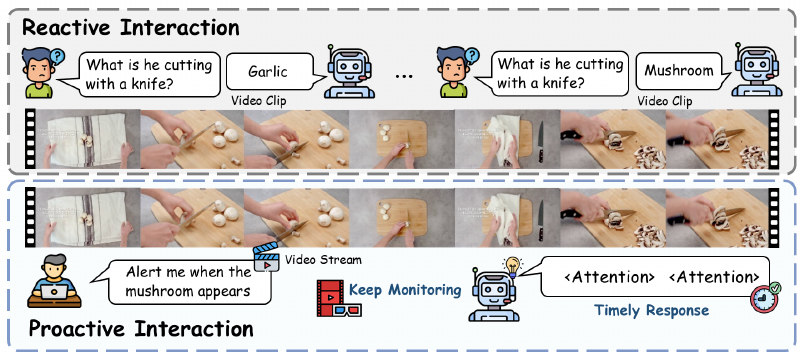} 
  \caption{Reactive Interaction v.s. Proactive Interaction paradigms}
  \label{fig:motivation}
\end{figure}

The rapid advancement of Multimodal Large Language Models
has revitalized the field of Human-Machine Interaction, catalyzing a wide array of intelligent applications that significantly enhance human productivity and reshape interactive experiences. 
Researchers have introduced numerous novel architectures and diverse training strategies, continuously expanding the capability of MLLMs ~\cite{xie2025mini, li2026qwen3}. 
However, prevalent MLLMs remain largely confined to a ``reactive response" paradigm~\cite{yang2025proagent}, where models generate responses solely upon receiving human instructions. 
Consequently, these models lack the capacity for proactive perception, contextual integration, and spontaneous interaction. 
As illustrated in Fig. \ref{fig:motivation}, proactive interaction necessitates continuous monitoring of streaming inputs and autonomous response timing grounded in visual and user contexts \cite{deng2025proactive}. 
The passive nature of existing MLLMs thus hinders their effective deployment in complex real-world settings.

Benchmarks are pivotal for tracking the rapid advancements in this domain. 
Extensive research \cite{tang2025video, kumar2025videollm} has been dedicated to evaluating the visual understanding capabilities of MLLMs (e.g., VideoMME \cite{fu2025video}, VideoMMMU \cite{hu2025video}, SVBench \cite{yang2025svbench}, and MVBench \cite{li2024mvbench}). 
However, mainstream benchmarks predominantly rely on a reactive QA paradigm that depends on explicit instructions, failing to quantify proactive capabilities such as autonomous perception or dialogue initiation.
This creates an evaluation bias: models that claim proactivity are still assessed using passive metrics, which fail to reflect their actual real-world performance. \cite{chen2024videollm, zhang2025eyes}. 
To address this limitation, works such as OmniMMI \cite{wang2025omnimmi} and StreamingBench \cite{lin2024streamingbench} introduced ``active alerting" tasks driven by emergent visual objects. 
These benchmarks lay the groundwork for proactive evaluation, marking a transition from reactive response to proactive perception.

However, current benchmarks exhibit limitations in that they treat proactive evaluation as an auxiliary subset, resulting in limited scale and insufficient fine-grained annotations.
Furthermore, the absence of unified evaluation standards leads to inconsistent protocols that fail to accurately quantify response timeliness and quality, severely hindering effective assessment. Ultimately, current benchmarks primarily focus on generic active alerting, overlooking the personalized intent understanding essential for practical assistance (as shown in Table \ref{tab:benchmark_compare}). 
Consequently, they lack a systematic framework for evaluating model capabilities under complex proactive scenarios.
Existing works indicate that the construction of high-quality proactive interaction datasets faces fundamental challenges.
First, precise temporal alignment in video streams is critical. Current benchmarks often suffer from misalignment between actual visual events and temporal annotations, introducing noise that interferes with model reasoning.
Moreover, personalized assistants rely heavily on egocentric perspectives. 
Within such non-stationary environments, constant user movement and camera rotation lead to intermittent occlusion or re-emergence of visual targets, which are often inconspicuous or minute in scale.
Such visual instability significantly complicates the task of high-precision temporal annotation.
Finally, proactive agents are required to interpret user intent and deliver feedback at optimal timing. 
The complexity of simulating realistic user profiles, maintaining accurate user memory, and leveraging user priors for context-aware reminders constitutes a major hurdle in modeling real-world proactive interactions.

To address the aforementioned challenges, we propose \textbf{EgoPro-Bench}, a comprehensive benchmark tailored for personalized proactive interaction. 
As shown in Fig. \ref{fig:teaser}, {EgoPro-Bench} encompasses egocentric data across multiple domains, including active alerting, navigation for the visually impaired, and tourism, while integrating detailed user profiles. 
As shown in Table~\ref{tab:benchmark_compare}, EgoPro-Bench is designed to benchmark the proactive capabilities and personalized responsiveness of MLLMs. In terms of data construction, we employed a streaming processing pipeline that performs frame-by-frame analysis and executes rigorous filtering to ensure the validity of visual event objects and the precision of temporal annotations.
Regarding personalization simulation, we leveraged advanced MLLMs to generate diverse user attributes, producing authentic user characteristics and memories.
Based on these profiles, we customized user memory construction aligned with the visual content of each domain and generated corresponding proactive responses, with all intermediate results subjected to rigorous quality assurance.

Moreover, we present \textbf{ProAct-Stream}, a proactive streaming model that embodies the ``short thinking, better interaction" paradigm. 
By strictly limiting reasoning tokens before intent recognition, ProAct-Stream effectively balances the trade-off between low-latency responsiveness and interaction precision.
We further evaluate a wide range of MLLMs on EgoPro-Bench and existing datasets under a unified protocol. 
Experimental results show that while existing models possess initial proactive capabilities, they fail to accurately determine interaction timing or handle complex scenarios.
In contrast, our model achieves substantial performance improvements.
Notably, we observe that smaller models sometimes outperform larger ones, indicating that simply scaling up model parameters does not guarantee performance gains in proactive tasks.

Our contributions are summarized as follows:
\begin{itemize}
    \item We propose EgoPro-Bench, a comprehensive benchmark for personalized proactive interaction, which covers various egocentric domains with simulated user profiles to evaluate MLLMs in proactive scenarios.
    \item We introduce an active interaction paradigm, ``shorter thinking, better interaction", and empirically validate its effectiveness on the EgoPro-Bench training set, demonstrating improved proactive interaction capabilities of MLLMs under real-time interaction constraints.
    \item We establish a robust data pipeline that utilizes streaming annotation to enable precise and personalized contextual modeling, alongside a unified evaluation protocol that quantifies proactive capabilities via multi-dimensional metrics in a multi-turn streaming framework.
    \item We present the ProAct-Stream model and evaluate advanced MLLMs, revealing the inherent shortcomings of reactive paradigms in interaction timing and confirming the advantages of our proactive model.
\end{itemize}



\begin{table*}[ht]
\centering
\caption{Comparison of reactive and proactive video understanding benchmarks. MC: Multiple Choice. OE: Open-Ended. Streaming: Video is processed in a streaming (frame-by-frame) manner. Egocentric: First-person perspective. Proactive: Supports proactive interaction scenarios. Intent: Includes personalized intent understanding scenario. $\cmark$: Fully supported. $\tmark$: Partially supported. {*}: Only proactive interaction subset. }
\resizebox{0.95\linewidth}{!}{
\begin{tblr}{
  colspec = {l c c c c c c c},
  row{1} = {font=\bfseries}, 
  row{2} = {bg=blue!4}, 
  row{9} = {bg=green!4}, 
  row{14} = {bg=red!6, font=\bfseries}, 
  column{even} = {c},
  column{3} = {c},
  column{5} = {c},
  column{7} = {c},
  cell{2}{1} = {c=8}{},
  cell{9}{1} = {c=8}{},
  vline{2} = {1,3-8,10-14}{0.05em},
  hline{1,15} = {-}{0.08em},
  hline{2-3,9-10,14} = {-}{0.05em},
}
\textbf{Benchmark}  & \textbf{\# Videos} & \textbf{QA Type} & \textbf{Annotation} & \textbf{Streaming} & \textbf{Egocentric} & \textbf{Proactive} & \textbf{Intent} \\
\textbf{Reactive Benchmarks} &                    &                  &                     &                    &                     &                    &                 \\
EgoSchema \cite{mangalam2023egoschema}& 5030               & MC               & Manual              & $\xmark$                   & $\cmark $                  & $\xmark $                  & $\xmark$                \\
MVBench \cite{li2024mvbench}          & 3641               & MC               & Auto                & $\xmark$                   & $\tmark$                   & $\xmark$                   & $\xmark$                \\
VideoMME \cite{fu2025video}           & 900                & MC               & Manual              & $\xmark$                   & $\xmark$                    & $\xmark$                   & $\xmark$                \\
VideoMMMU \cite{hu2025video}          & 300                & MC               & Manual              & $\xmark$                   & $\xmark$                   & $\xmark$                   & $\xmark$                \\
SVBench \cite{yang2025svbench}        & 1353               & OE               & Auto \& Manual      & \cmark                  & \tmark                  & \xmark                   & \xmark                \\
MLVU \cite{zhou2025mlvu}              & 2593               & MC+OE            & Auto \& Manual      & $\xmark$                   & $\tmark$                  & $\xmark$                  & $\xmark$                \\
\textbf{Proactive Benchmarks}                  &                    &                  &                     &                    &                     &                    &                 \\
StreamingBench\textsuperscript{*}  \cite{lin2024streamingbench} & 50           & MC+OE            & Auto \& Manual      & $\cmark$                  & $\xmark$                   & $\tmark$                  & $\xmark$                \\
OmniMMI\textsuperscript{*} \cite{wang2025omnimmi}        & 200                & OE               & Auto \& Manual      & $\xmark$                  & $\xmark $                  & $\tmark$                  & $\xmark$                \\
OvO-Bench\textsuperscript{*}  \cite{niu2025ovo}           & 82                 & MC+OE            & Auto \& Manual      & $\cmark$                  & $\tmark $                  & $\tmark$                  & $\xmark$                \\
StreamGaze \cite{lee2025streamgaze}   & 285                & MC+OE            & Auto \& Manual      & $\cmark$                  & $\cmark$                   & $\tmark $                & $\xmark$                \\
EgoPro-Bench (Ours)                   & 2400               & OE               & Auto \& Manual      & $\cmark$                   & $\cmark$                   & $\cmark$                  & $\cmark $               
\end{tblr}
}
\label{tab:benchmark_compare}
\end{table*}

%


\section{Related Work}

\subsection{Video Understanding Benchmarks.}
Prior video understanding benchmarks have targeted diverse capability dimensions. Spatial and temporal perception are evaluated by VSI-Bench~\cite{yang2025thinking} and MVBench~\cite{li2024mvbench}, respectively, while VideoMMMU~\cite{hu2025video} focuses on multimodal knowledge. For long-horizon comprehension, LVBench~\cite{wang2025lvbench} serves as a key standard. Furthermore, benchmarks like MMVU~\cite{zhao2025mmvu} and VideoMME~\cite{fu2025video} specifically assess complex multi-step reasoning.

\subsection{Streaming Video Benchmarks.}
Recent work has proposed several benchmarks for streaming video understanding, aiming to evaluate models in more interactive settings \cite{fu2025vispeak,yang2025svbench}.  StreamingBench~\cite{lin2024streamingbench}, and OvOBench~\cite{niu2025ovo} process videos incrementally and assess models through queries about the video context at intermediate points, leading to predominantly reactive, query-driven interactions. StreamGaze~\cite{lee2025streamgaze} further incorporates human gaze signals to facilitate temporal reasoning in streaming videos. However, 
current methods predominantly rely on external triggers or predefined QA tasks. To address this, we introduce EgoPro-Bench, which shifts the paradigm to intent-driven interaction. By incorporating efficient short thinking \cite{hassid2025don, wang2025thinking} for real-time responsiveness, while uniquely conditioning on user memory and dynamic scene context, our framework operates without explicit external instructions.

\begin{figure*}[ht]
  \centering
  \includegraphics[
    width=\textwidth,
    height=0.28\textheight,
  ]{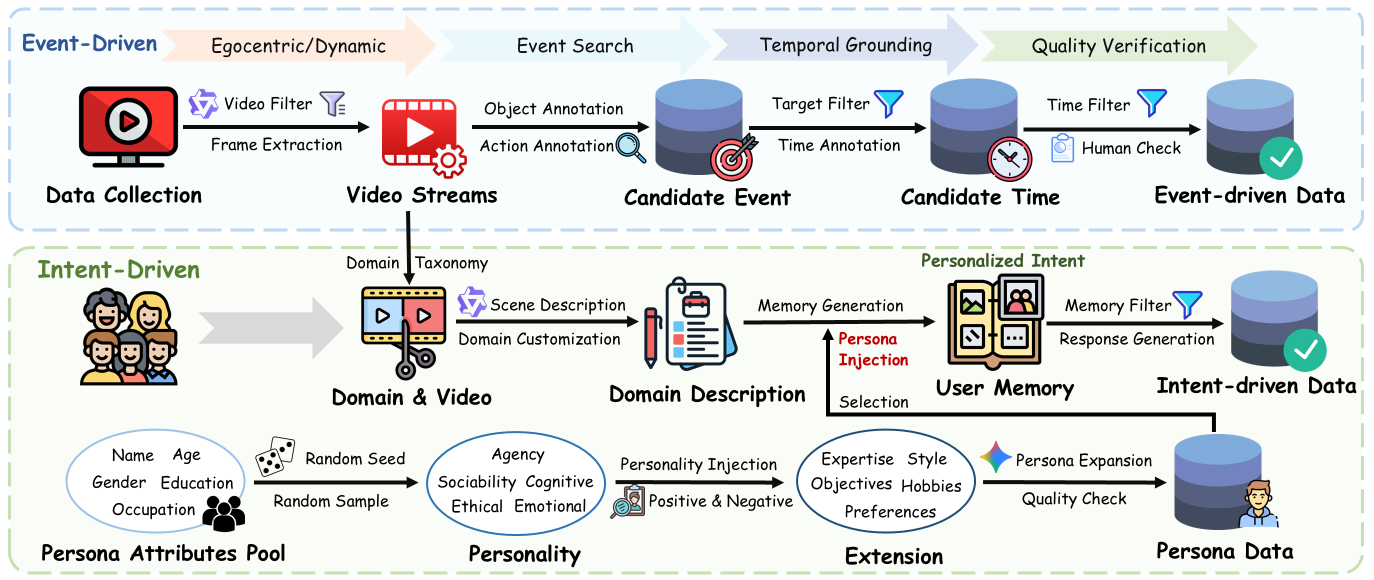} 
  \caption{Data synthesis pipeline for event-driven and intent-driven proactive interaction. The event-driven branch divides tasks into ``object" and ``action", focusing on visual and temporal precision. The intent-driven branch synthesizes personalized user intents by injecting persona profiles into diverse domain scenarios. Strict data filtering and quality checks are applied throughout the pipeline.}
  \label{fig:pipeline}
\end{figure*}

\section{Benchmark}
\subsection{Overview}
We propose EgoPro-Bench, the pioneering benchmark for personalized proactive MLLMs, which include event-driven and intent-driven branches.
It adopts an egocentric perspective across scenarios ranging from event alerting to daily assistance. As shown in Figure~\ref{fig:pipeline}, to guaranty high-quality data, our construction pipeline synergizes streaming analysis with user profiling, generating realistic and personalized interaction contexts. Through this systematic process, we ultimately curated 2,400 personalized videos for the benchmark, significantly expanding the scope of proactive interaction evaluation. This positions EgoPro-Bench as a foundational bridge from reactive paradigms to next-generation user-centric agents.

\subsection{Data Collection}
To construct a personalized proactive interaction benchmark, as shown in Table~\ref{tab:data_collection_used}, we collect egocentric video data from open datasets, ensuring the diversity and fidelity essential for realistic simulations.

\subsection{Data Domain}
To systematically assess proactive interaction capabilities in diverse scenarios, we structure 12 core domains for event-driven and intent-driven branches. 
The event-driven branch focuses on proactive alerting tasks and is further categorized into \textit{object} and \textit{action} tasks, corresponding to the \textit{object} and \textit{action} domains, respectively.
The former emphasizes the identification and alerting of specific visual objects, while the latter prioritizes the monitoring of scene events or human actions. 
The intent-driven branch encompasses 10 common egocentric living scenarios: \textit{working}, \textit{travel}, \textit{sports}, \textit{art}, \textit{navigation}, \textit{dailylife}, \textit{shopping}, \textit{cooking}, \textit{driving}, and \textit{entertainment}. Specifically, the navigation domain is tailored to assess obstacle avoidance for the visually impaired. 





\begin{table}[ht]
\caption{Statistics of video data collection from each dataset and the final data used in EgoPro-Bench.}
\centering
\small
\begin{tabular}{ccc}
\hline
\textbf{Dataset} & \textbf{Collection} & \textbf{Used}  \\
\hline
EgoBlind \cite{xiao2025egoblind}        & 1259         & 420               \\
StreamGaze \cite{lee2025streamgaze}     & 230          & 68            \\
EgoExoLearn \cite{huang2024egoexolearn} & 8492         & 387            \\
EgoTextVQA \cite{zhou2025egotextvqa}    & 2561         & 833             \\
Egoschema \cite{mangalam2023egoschema}  & 29650        & 5411               \\
LLaVA-Video \cite{zhang2024video}       & 12822        & 4075                 \\
Ego4D \cite{grauman2022ego4d}           & 14563        & 2289                    \\
EgoQA \cite{nguyen2024encoding}         & 7768         & 856                  \\
\hline
\textbf{Sum}         &  \textbf{77345}         & \textbf{14339}        \\
\hline
\end{tabular}
\label{tab:data_collection_used}
\end{table}

\subsection{Benchmark Construction}



\subsubsection{Data Preprocessing.}
We sampled raw videos at 1.0 FPS and leveraged Qwen3-VL-30B-A3B-Instruct \cite{bai2025qwen3vltechnicalreport} to isolate high-dynamic egocentric sequences suitable for proactive interaction, strictly excluding static or information-sparse content. 

\textbf{Object Annotation.} The construction of event-driven data begins with identifying key objects and scene events. We utilize MLLMs to extract candidate events from video streams, enforcing a strict filtering criterion: target objects must exhibit transient visibility rather than persistent presence (e.g., background elements). This constraint ensures that interaction tasks possess distinct temporal triggers. To guarantee data quality, we implement a secondary verification mechanism where MLLMs assess annotation accuracy and logical consistency, filtering out low-quality samples. Detailed prompt designs are provided in Appendix~\ref{Object_Annotation}.
\subsubsection{Event-driven Pipeline.}


\textbf{Time Annotation.} To address the prevalent issue of temporal annotation bias in proactive scenarios, particularly the intermittent occlusion and re-emergence of visual targets caused by rapid motion in egocentric views data, we employed a streaming frame-by-frame annotation strategy. This strategy employs MLLMs for frame-wise object inference, reinforced by rigorous automated filtering and human verification to mitigate noise. This pipeline effectively resolves temporal localization challenges in complex scenes, ensuring high-precision timestamps (detailed prompts provided in the Appendix~\ref{Time_Annotation}).

\subsubsection{Intent-driven Pipeline.}
\textbf{User Profile Synthesis.} 
To simulate the complex intent and memory mechanisms of real-world users, we designed a hierarchical persona synthesis pipeline inspired by Persona-hub \cite{ge2024scaling}. 
First, we initialized foundational prototypes by sampling from a comprehensive attribute pool.
To enhance behavioral realism, we injected personality priors across dimensions such as agency, sociability, morality, and emotion, incorporating both positive and negative biases to ensure heterogeneity. 
Finally, we leveraged MLLMs to expand profiles with fine-grained details, including preferences and expertise, under rigorous quality control. 
This strategy yielded a diverse collection of realistic personas covering more than 500 professions and 300 personality descriptors, detailed profiles are provided in Appendix \ref{persona_profile_details}.


\textbf{Domain Customization.} 
Initially, we categorized video streams via a domain taxonomy. Leveraging domain-specific prompts, we employed MLLMs to generate detailed domain and visual descriptions that capture the scene's core semantics and latent interaction needs. These fine-grained descriptions serve as essential conditional priors for memory synthesis and response generation, enhancing perceptual precision.




\begin{table}[ht]
\caption{Statistics of video data across different domains in the training and test sets.}
\centering
\small
\begin{tabular}{cccc}
\hline
\textbf{Domain} & \textbf{SFT} & \textbf{RL} & \textbf{Test} \\
\hline
Object           & 1196         & 300         & 200           \\
Action            & 1199         & 300         & 200           \\
Cooking          & 990          & 154         & 200           \\
Dailylife        & 4340         & 665         & 200           \\
Driving          & 166          & 26          & 200           \\
Entertainment    & 401          & 58          & 200           \\
Navigation       & 195          & 34          & 200           \\
Art              & 117          & 17          & 200           \\
Shopping         & 449          & 58          & 200           \\
Sports           & 313          & 50          & 200           \\
Travel           & 89           & 21          & 200           \\
Working          & 830          & 117         & 200  \\
\hline
\textbf{Sum}         &  \textbf{10285}         & \textbf{1800}         & \textbf{2400}   \\
\hline
\end{tabular}
\label{tab:data_distribution}
\end{table}
\textbf{Memory Generation.} 
Synthesizing domain descriptions, video semantics, and user profiles, we generate user memories to simulate long-term interaction contexts. This process aims to construct authentic intent understanding scenarios. 
To address domain disparities, we employ a domain-adaptive strategy that aligns memory generation with scene-specific constraints. In safety-critical scenarios (e.g., navigation for the visually impaired), the system prioritizes stable preferences regarding behavioral habits and environmental adaptation. In contrast, for daily life and entertainment contexts, the focus shifts to capturing nuanced user interests and personalized needs. To ensure data fidelity, we implement a rigorous dual-filtering mechanism based on content plausibility and visual relevance, eliminating inconsistencies to yield a robust memory corpus.

\textbf{Response Generation.} In the response generation phase, we construct proactive interventions at precise moments, ensuring dual consistency between personalized user history and the visual scene. By conditioning on user memory and domain context, the model autonomously determines optimal intervention timing and content. To ensure quality, we employ an LLM-based filtering mechanism to retain only high-confidence responses, yielding a robust intent-driven dataset. Detailed prompts are provided in Appendix \ref{sec:response_generate}.

\subsection{Data Statistics}
The statistical distribution of video data across various domains is presented in Table \ref{tab:data_distribution}. Our training set encompasses over 12,000 samples, ensuring extensive data richness and diversity. We also established a balanced test benchmark, featuring 200 samples per domain. In terms of annotation strategy, Supervised Fine-Tuning (SFT) samples are paired with a concise Chain-of-Thought (CoT) to bolster reasoning, while  Reinforcement Learning (RL) data adopts a streaming multi-turn dialogue structure tailored to optimize the model using historical context and the final response. Notably, distinct from event-driven data, all intent samples are explicitly grounded in specific user memory profiles.





\section{Method}
We introduce ProAct-Stream, a proactive interaction model engineered for streaming video understanding that encapsulates the ``short thinking, better interaction" paradigm. By employing a two-stage training paradigm that integrates SFT and RL, we empower the model with efficient reasoning and low-latency response characteristics. This approach facilitates accurate and timely interactions within complex streaming environments.
\subsection{Stage-1: SFT}
We initiate the training phase with full-parameter SFT on CoT data, optimizing the cross-entropy objective. Diverging from standard CoT paradigms, our dataset is tailored for streaming multi-turn contexts. To guarantee low-latency responses, we enforce a constraint for concise yet insightful reasoning, compelling the model to achieve accurate visual assessment with minimal token overhead. Specific CoT generation templates are detailed in Appendix \ref{appenix_cot}.

\subsection{Stage-2: RL} 
We leverage the Group Sequence Policy Optimization (GSPO)~\cite{zheng2025groupsequencepolicyoptimization} to conduct RL training based on LoRA. Aiming to drive better interaction via minimal thinking overhead, we implement a joint optimization strategy focusing on both reasoning length and response quality. To this end, for two sentences $s_1$ and $s_2$, we define two distinct functions, textual similarity and semantic similarity: 
\begin{align}
sim_{text}(s_1, s_2) &= \frac{2 \cdot M}{L_1 + L_2} \\
sim_{sem}(s_1, s_2) &= \frac{LLM(s_1, s_2) - 1}{3} 
\end{align}
where $M$ denotes the number of matched characters and $L$ represents the length of the string. Semantic similarity is evaluated by an LLM to assign a score on a scale of 1 to 4. 

\subsubsection{Proactive Reasoning Reward}
We formulate the proactive reasoning reward, comprising four components:

\textbf{Format Reward.} We employ a widely adopted format reward $R_{format}$ to verify that the model's reasoning is correctly enclosed within the {\textit{\textless think\textgreater\textless/think\textgreater}} tags. We assign a format reward of 1 if the output adheres to the prescribed format. Otherwise, the entire reasoning reward is penalized to 0.

\textbf{Length Reward.} To avoid slow interaction caused by excessive reasoning, we employ a length reward with linear decay to regulate the output length.  Let $L$ denote the reasoning length; the reward linearly decreases to zero within the interval $[L_{\min}, L_{\max}]$, where we set $L_{\min}=16$, $L_{\max}=22$ by default. The reward is formulated as follows:
\begin{align}
R_{len}=1-\max\left(0,\min\left(1,\frac{L-L_{min}}{L_{max}-L_{min}}\right)\right)
\end{align}
\textbf{Historical Diversity Reward.} This reward is designed to penalize content that is overly similar to past reasoning, thereby encouraging the model to generate more diverse thoughts. Let $H=\{h_1, h_2, \dots, h_n\}$ denote the set of historical reasoning segments, and let $s_{\text{th}}$ represent the current reasoning content. The reward is defined as follows:
\begin{align}
R_{hist}=1-\max_{h\in H}\{sim_{text}(s_{th},h)\}
\end{align}
\textbf{Semantic Consistency Reward.} We utilize the semantic consistency reward $R_{sem}$ to refine the generated reasoning. This mechanism assesses both the textual and semantic alignment against ground-truth reasoning $g_{th}$ to guarantee the validity of the reasoning. The definition is given by:
\begin{align}
R_{sem}=\frac{1}{2}\left(sim_{text} (s_{th},g_{th})+sim_{sem}(s_{th},g_{th})\right)
\end{align}
We aggregate all individual reward components to derive the final proactive reasoning reward function $R_{reason}$:
\begin{align}
R_{reason}=R_{format} + R_{len} + R_{hist} + R_{sem}
\end{align}
\subsubsection{Proactive Response Reward}
We optimize the final response using a proactive reward that accounts for both length and content quality.

\textbf{Response Length Reward.}
The length reward follows the same mechanism as its reasoning counterpart but utilizes a narrower boundary range to encourage concise responses. And, we set $L_{\min}=26$, $L_{\max}=37$ by default.

\textbf{Response Content Reward.} We predefine a response set $G_{d} = \{\textit{\textless Attention\textgreater}, \textit{\textless Silence\textgreater}\}$. If the ground truth $g_{r}$ corresponds to a special symbol within this set, we employ solely textual similarity to assess the model response $c_{r}$. Otherwise, we combine textual and semantic similarities to formulate the reward signal:
\begin{align}
R_{cont}=
\begin{cases}
sim_{text}(c_{r},g_{r}) & \mathrm{if}\ g_{r}\in G_{d} \\
sim_{text}(c_{r},g_{r})+sim_{sem}(c_{r},g_{r}) & \mathrm{otherwise}  \\
\end{cases}
\end{align}
Consequently, we derive the final proactive response reward and the joint optimization reward as follows:
\begin{align}
& R_{resp}=R_{len} + R_{cont} \\
& R=R_{reason} + R_{resp} 
\end{align}
We utilize this reward signal to drive the GSPO algorithm, aiming to achieve a rapid and smart proactive interaction experience in streaming scenarios through concise reasoning.





\section{Experiments}

\begin{table*}[ht]
\centering
\caption{Performance comparison on three public benchmarks. \textbf{Bold} indicates the best performance, while \underline{underlined} denotes the second-best result. SFT refers to Qwen3VL-8B, fine-tuned on the EgoPro-Bench training set.}
\resizebox{\linewidth}{!}{
\begin{tblr}{
  colspec = {l *{15}{c}},
  cell{1}{1} = {r=2}{c}, 
  cell{1}{2} = {c=5}{c}, 
  cell{1}{7} = {c=5}{c}, 
  cell{1}{12} = {c=5}{c}, 
  row{3} = {bg=red!5},   
  row{14} = {bg=blue!5},  
  hline{1,16} = {-}{0.08em}, 
  hline{2} = {2-16}{0.03em}, 
  hline{3,14,16} = {-}{0.05em},
  vline{7,12} = {-}{0.05em},
}
\textbf{Model} & \textbf{OmniMMI} & & & & & \textbf{OvO-Bench} & & & & & \textbf{StreamingBench} & & & & \\
  & Precision & Recall & F1. & GHA & mIoU &  Precision & Recall & F1. & GHA & mIoU & Precision & Recall & F1. & GHA & mIoU  \\
\textbf{Open-Source Models} & & & & & & & & & & & & & & & \\
VideoChat-R1.5  & 33.01 & 44.48 & 32.71 & 78.25 & 26.47  & 38.43 & 44.05 & 32.58 & 68.65 & 24.80  & \textbf{41.73} & 27.16 & 24.28 & 82.67 & 15.70 \\
VideoRFT-7B  & 6.86 & 33.32 & 10.32 & 63.00 & 6.58  & 31.02 & 67.01 & 37.76 & 56.45 & 29.13  & 31.06 & 40.40 & 20.34 & 74.93 & 13.29 \\
Qwen2.5vl-3B  & 1.72 & 5.05 & 2.15 & 64.17 & 1.49  & 1.16 & 7.07 & 1.95 & 57.60 & 1.15 &  2.78 & 4.40 & 1.76 & 67.07 & 3.49 \\
Qwen2.5VL-7B  & 14.06 & 30.50 & 16.29 & 68.67 & 12.47 & 28.87 & 41.42 & 28.84 & 66.33 & 22.77  & 33.65 & 29.60 & 20.04 & 76.00 & 13.77 \\
Qwen2.5VL-32B  & 34.99 & 20.89 & 23.65 & 76.50 & 19.35  & 26.67 & 20.32 & 20.21 & 69.08 & 15.85  & 35.93 & 25.32 & 23.68 & 80.80 & 16.76 \\
Qwen2.5VL-72B  & \underline{62.13} & 51.45 & 50.93 & 90.25 & 42.22  & \underline{54.18} & \underline{51.87} & \underline{43.95} & 79.30 & \underline{34.43}  & 39.49 & 36.52 & 31.06 & \underline{84.80} & \underline{22.02} \\
Qwen3VL-4B  & 54.25 & \underline{70.81} & \underline{55.94} & \underline{90.50} & 45.83 & 47.05 & 45.83 & 41.13 & 80.96 & 30.87  & 30.42 & \underline{50.80} & \underline{31.81} & 77.47 & 21.60 \\
Qwen3VL-30B-A3B  & 39.77 & 28.99 & 30.37 & 79.58 & 24.97 &  \textbf{58.77} & 29.30 & 34.02 & \textbf{84.76} & 24.55  & 40.14 & 32.12 & 31.13 & \textbf{88.27} & 21.47 \\
Qwen3VL-8b  & 58.12 & 59.58 & 52.82 & 89.67 & 44.43  & 38.61 & 32.88 & 26.59 & 71.20 & 19.77 & \underline{41.02} & 38.76 & 31.49 & 83.83 & 21.07 \\
TimeChatOnline-7B  & 3.89 & 15.25 & 5.60 & 65.08 & 3.77  & 9.89 & 17.38 & 10.47 & 59.58 & 7.82  & 33.06 & 13.32 & 12.84 & 78.00 & 9.30 \\

\textbf{ProAct-Stream (Ours)} & & & & & & & & & & & & & & & \\
SFT  & \textbf{65.24} & \textbf{71.94} & \textbf{62.85} & \textbf{96.92} & \textbf{51.05}  & 50.01 & \textbf{76.30} & \textbf{52.98} & \underline{82.24} & \textbf{41.01} & 34.14 & \textbf{54.32} & \textbf{35.43} & 83.62 & \textbf{25.11} \\
\end{tblr}
}
\label{tab:event_base_result}
\end{table*}

\begin{table*}[ht]
\centering
\caption{Performance comparison on three subsets of EgoPro-Bench. \textbf{Bold} indicates the best performance, \underline{underlined} denotes the second-best result, while {*} means the best result in our ProAct-Stream models. MC denotes Memory Consistency, and RQ represents Response Quality. SFT denotes Qwen3VL-8B fine-tuned on EgoPro-Bench, while RL indicates further training from the SFT checkpoint. W and W/O Think denote whether \textit{think} data are used during SFT. The full domain's results of EgoPro-Intent are reported in Appendix~\ref{intent_driven_detals_result}.}
\resizebox{\linewidth}{!}{
\begin{tblr}{
  colspec = {l *{19}{c}},
  cell{1}{1} = {r=2}{c}, 
  cell{1}{2} = {c=5}{c}, 
  cell{1}{7} = {c=5}{c}, 
  cell{1}{12} = {c=7}{c}, 
  row{3} = {bg=red!5},   
  row{14} = {bg=green!5},
  hline{1,19} = {-}{0.08em}, 
  hline{2} = {2-22}{0.03em}, 
  hline{3,14, 18} = {-}{0.05em}, 
  vline{7,12} = {-}{0.05em},
}
\textbf{Model} & \textbf{EgoPro-{Action}} & & & & & \textbf{EgoPro-Object} & & & & & \textbf{EgoPro-Intent} & & & & & & \\
  & Precision & Recall & F1. & GHA & mIoU  & Precision & Recall & F1. & GHA & mIoU  & Precision & Recall & F1. & GHA & mIoU & MC & RQ \\
\textbf{Open-Source Models} & & & & & & & & & & & & & & & & & \\

VideoChat-R1.5 &    55.73 & 37.17 & 37.18 & 62.50 & 30.74 &
                    56.80 & 42.06 & 40.93 & 65.20 & 33.87 & 
                    8.19 & 76.09 & 13.82 & 60.54 & 54.44 & 3.34 & 1.62 \\
VideoRFT-7B &       36.16 & 41.90 & 34.19 & 55.36 & 29.78 & 
                    43.64 & 61.75 & 44.15 & 55.40 & 35.52 & 
                    8.64 & 95.11 & 15.29 & 54.21 & 65.66 & \underline{4.19} & 1.91 \\
Qwen2.5vl-3B &      6.31 & 5.88 & 4.73 & 50.86 & 4.06 & 
                    5.33 & 4.02 & 3.67 & 52.49 & 2.69 & 
                    8.91 & \textbf{100.00} & 15.80 & 52.70 & \underline{68.71} & 3.90 & 1.68 \\
Qwen2.5VL-7B &      40.72 & 44.25 & 36.45 & 57.18 & 30.29 & 
                    40.72 & 44.25 & 36.45 & 58.91 & 30.29 & 
                    8.76 & 97.00 & 15.48 & 54.09 & 66.75 & 4.06 & 2.10 \\
Qwen2.5VL-32B &     45.53 & 11.17 & 14.67 & 59.58 & 10.52 & 
                    68.93 & 48.52 & 52.97 & 73.35 & 46.50 & 
                    8.95 & \underline{99.54} & 15.84 & 55.96 & 68.49 & 4.13 & 2.18 \\
Qwen2.5VL-72B &     67.90 & 28.58 & 33.72 & 64.10 & 25.48 & 
                    90.67 & 83.17 & 84.01 & 83.03 & 77.37 & 
                    9.34 & 98.86 & 16.14 & 57.39 & \textbf{68.89} & 3.95 & 1.98 \\
Qwen3VL-4B &        74.76 & 48.65 & 52.24 & 66.63 & 42.13 & 
                    79.11 & 77.61 & 75.01 & 79.54 & 67.70 & 
                    13.29 & 73.09 & 18.04 & 68.33 & 55.19 & 3.49 & 2.10 \\
Qwen3VL-30B-A3B &   84.99 & 42.36 & 49.47 & 75.21 & 38.91 & 
                    \underline{90.97} & 54.75 & 62.28 & 82.51 & 53.24 & 
                    10.13 & 90.42 & 16.88 & 66.53 & 65.88 & \textbf{4.21} & 2.67 \\
Qwen3VL-8b  &       72.90 & 28.85 & 34.60 & 66.40 & 26.72 & 
                    59.45 & 47.96 & 49.35 & 71.85 & 44.57 & 
                    15.30 & 76.50 & 19.94 & 68.68 & 57.21 & 3.87 & 2.35 \\
TimeChatOnline-7B & 30.32 & 41.98 & 33.12 & 52.62 & 29.05 & 
                    18.83 & 22.83 & 18.56 & 55.16 & 15.48 & 
                    7.48 & 80.59 & 13.11 & 54.05 & 55.40 & 3.46 & 1.60 \\

\textbf{ProAct-Stream (Ours)} & & & & & & & & & & & & & & & & & \\
SFT &               \underline{88.64} & \underline{84.74} & \underline{84.78} & \textbf{83.00}\textsuperscript{*} & \underline{78.54} & 
                    90.78 & 87.24 & 88.00 & \underline{88.07} & \underline{84.01} & 
                    \underline{71.75} & 50.07 & 55.02 & 72.34 & 56.28 & 3.81 & \underline{3.02} \\
RL W/O Think &      \textbf{88.86}\textsuperscript{*} & \textbf{85.10}\textsuperscript{*} & \textbf{85.03}\textsuperscript{*} & \underline{82.79} & \textbf{78.66}\textsuperscript{*} & 
                    90.78 & \underline{87.36} & \underline{88.07} & 87.94 & 84.12 & 
                    \textbf{72.15}\textsuperscript{*} & 50.11 & \underline{55.14} & \underline{72.39} & 55.96 & 3.81 & 3.01 \\
RL W Think&         82.93 & 81.52 & 79.66 & 80.04 & 72.92 & 
                    \textbf{94.00}\textsuperscript{*} & \textbf{90.22}\textsuperscript{*} & \textbf{91.19}\textsuperscript{*} & \textbf{89.89}\textsuperscript{*} & \textbf{87.01}\textsuperscript{*} & 
                    66.50 & 57.86\textsuperscript{*} & \textbf{56.34}\textsuperscript{*} & \textbf{76.13}\textsuperscript{*} & 61.62\textsuperscript{*} & 4.10\textsuperscript{*} & \textbf{3.23}\textsuperscript{*} \\

\end{tblr}
}
\label{tab:intent_based_result}
\end{table*}

\subsection{Experimental Setup}
\textbf{Baselines.} To conduct a comprehensive and available evaluation, we selected open-source MLLMs, encompassing the Qwen2.5VL/Qwen3VL-Instruct series~\cite{bai2025qwen2, bai2025qwen3vltechnicalreport}, VideoChat-R1.5\cite{yan2025videochat}, VideoRFT \cite{wang2025videorft}, and the TimeChat-Online~\cite{yao2025timechat} streaming models. 



\textbf{Benchmarks.} To validate generalization ability in proactive interaction scenarios, we selected the proactive alerting subsets from OmniMMI \cite{wang2025omnimmi}, StreamingBench \cite{lin2024streamingbench}, and OvOBench \cite{niu2025ovo} as external benchmarks, conducting standardized tests under our proposed unified evaluation protocol.  Regarding our proposed EgoPro-Bench, we subdivided it into three core subsets, namely \textit{Object}, \textit{Action}, and \textit{Intent}, designed to quantitatively evaluate key competencies in object recognition, action perception, and intent understanding, respectively. 

\textbf{Implement Details.} To simulate authentic interaction scenarios, we implemented a streaming frame-by-frame evaluation protocol incorporating historical context. For event-driven tasks, we introduced two response patterns: \textit{\textless Attention\textgreater} to trigger a response and \textit{\textless Silence\textgreater} to maintain silence. Intent-driven tasks build upon this by leveraging user memory for autonomous responses, adhering to the same silence protocol. We developed ProAct-Stream based on the Qwen3VL-8B-Instruct, utilizing 16 NVIDIA H100 GPUs for training.
The training environments and the corresponding hyperparameter settings for both RL and SFT are fully presented in Appendix~\ref{experiment_setting}.


\subsection{Evaluation Protocol}
To more comprehensively evaluate the timing, frequency, and quality of model responses, we introduce objective metrics and LLM-based judge scores for performance assessment. The objective metrics primarily measure the timing and quantity of model responses, while the LLM judge scores focus on evaluating the overall quality of the generated response texts.

\textbf{Objective Metrics.} We adopt five objective evaluation metrics: Precision, Recall, F1-score, mean Intersection-over-Union (mIoU), and Ground-truth Hit Accuracy (GHA).
The GHA denotes the ratio of ground-truth (GT) intervals that contain at least one correct response, relative to the total number of GT intervals. Moreover, in intent-driven datasets, the model is expected to respond only once within each GT interval, in contrast to event-driven datasets, where alert signals are continuously issued throughout the entire GT interval. Therefore, in intent-driven evaluation, the definitions and computation of True Positive (TP), False Positive (FP), True Negative (TN), and False Negative (FN) differ from those used in event-driven settings. 



In particular, for intent-driven model evaluation, we employ the Hungarian algorithm to match model responses with GT intervals, enabling a more appropriate and reliable performance assessment.
Detailed procedures for computing these objective metrics are provided in Appendix \ref{objective_metrics_computation}.


\textbf{LLM-as-a-Judge.} To assess the quality of model responses in intent understanding tasks, we introduced two LLM-based evaluation metrics: \textbf{Memory Consistency} and \textbf{Response Quality}. Memory Consistency measures the alignment between the model's response and the user's long-term memory (e.g., preferences and background). Response Quality focuses on the effectiveness of the content, evaluating whether the response provides actionable alerts or practical suggestions. We utilize a 1-to-5 scoring scale. The evaluation adheres to a streaming frame-by-frame protocol utilizing a local context window. Detailed evaluation prompt templates are provided in the Appendix \ref{llm_judge}.



\subsection{Main Results and Analysis}
\subsubsection{Proactive Alerting Subsets Result}
As shown in Table~\ref{tab:event_base_result}, the results on existing public datasets demonstrate that model performance improves with increasing parameter scale. This observation indicates that current visual models possess strong capabilities to effectively understand objects and actions in videos and to generate accurate responses accordingly.  

Moreover, benefiting from the data provided by EgoPro-Bench, the ProAct-Stream model achieves state-of-the-art (SOTA) performance on the majority of evaluation metrics with only 8B parameters, and in some cases even outperforms models with 32B and 72B parameters.


\subsubsection{EgoPro-Bench Result} 
\textbf{Event-driven Result.}  The results of \textit{EgoPro-Action} and \textit{EgoPro-Object} in Table~\ref{tab:intent_based_result} show that open-source models achieve lower scores than our model, highlighting the challenges of event-driven streaming tasks and demonstrating that our data effectively improves the streaming response capabilities of the models. Additionally, we observed that compared to objects, actions suffer from visual instability and blurred boundaries. Such low observability makes the model prone to over-reasoning when visual anchors are missing, hindering precise judgments of interaction timing.

\textbf{Intent-driven Result.} Results for \textit{EgoPro-Intent} in Table~\ref{tab:intent_based_result} reveal that although baselines achieve higher Recall, their low Precision indicates a tendency to over-respond and violate silence constraints. This excessive reactivity leads to ill-timed interactions and compromised Response Quality. In contrast, our method attains higher Precision, showing that it learns to remain silent when interaction is unnecessary, which improves overall interaction quality despite slightly lower Recall.


Overall, as shown in Table~\ref{tab:intent_based_result}, existing MLLMs exhibit suboptimal performance across all EgoPro-Bench subsets, underscoring the benchmark’s rigor and the limitations of current reactive paradigms. Specifically, baseline models struggle to accurately capture interaction timing and align interaction content. In contrast, ProAct-Stream achieves a substantial performance improvement through the ``short thinking, better interaction'' paradigm. The ablation analysis reveals a clear performance trajectory: the introduction of concise reasoning during SFT enhances intent alignment, while the subsequent RL stage further maximizes response quality. These results show that allocating a limited amount of structured ``thinking'' is key to enabling more effective and proactive interaction, and demonstrate ``short thinking" can get a ``better interaction" performance.

\section{Conclusion}
In this paper, we presented \textbf{EgoPro-Bench}, a benchmark designed to bridge the gap between reactive MLLMs and personalized proactive agents. By synthesizing rigorous streaming pipelines with authentic user profiles, we address the critical lack of evaluation standards for interaction timing and personalization. Furthermore, we established a unified evaluation protocol and introduced \textbf{ProAct-Stream}, a proactive model optimized for streaming scenarios.
In intent-driven proactive interaction scenarios, we introduce the principle of ``short thinking, better interaction" and employ reinforcement learning to improve the model’s ability to perceive and infer user intent.
Our experiments reveal a pivotal insight: while current MLLMs possess strong perception, they struggle with the precise timing and intent alignment required for proactivity. 
We hope this work takes one step forward toward evolving MLLMs from reactive executors to proactive guardians integrated into daily life.




\section*{Impact Statement}


This paper presents work whose goal is to advance the field of Machine
Learning. There are many potential societal consequences of our work, none
which we feel must be specifically highlighted here.


\nocite{langley00}

\bibliography{reference}
\bibliographystyle{icml2026}

\newpage
\appendix
\onecolumn


\section{Prompt}
\subsection{Object Annotation}
\label{Object_Annotation}

\begin{tcolorbox}[
    title=\textbf{System Prompt for Emergent Object Annotation},
    colback=gray!5,
    colframe=gray!60!black,
    breakable,
    boxrule=0.8pt,
    fonttitle=\bfseries,
    left=1em, right=1em, top=1em, bottom=1em
]
\raggedright

You are a Video Analyst for a Smart Alert System. \\
Your task is to identify the single most significant ``Emergent Object" in the video.

\medskip
\# \textbf{THE PROBLEM} \\
We want to alert the user about \textbf{new} or \textbf{distinct} items that appear in the scene. \\
We don't want to find objects that persist in the video.

\medskip
\# \textbf{SELECTION LOGIC (Follow Strict Priority)}

\smallskip
1. \textbf{True Alerts (The ``Emergence" Rule):}
\begin{itemize}[label=*, leftmargin=1.5em, nosep]
    \item New Entries: An object that enters the frame (e.g., a car drives in, a bowl slides into view).
    \item Revealed Items: An object that is pulled out, picked up, or placed down (e.g., a person pulls a knife from a pocket, places a laptop on a table).
    \item Distinct Static Objects: An object sitting independently in the scene that is clearly the subject of the camera's focus (e.g., a package on a porch).
\end{itemize}

\smallskip
2. \textbf{INVALID Targets (Do NOT Select):}
\begin{itemize}[label=*, leftmargin=1.5em, nosep]
    \item Constant Accessories: An object that remains constant throughout the video.
    \item Background Clutter: Furniture or walls that never change state.
\end{itemize}

\medskip
\# \textbf{INSTRUCTIONS}
\begin{itemize}[label=-, leftmargin=1.5em, nosep]
    \item Watch the video. 
    \item Ask yourself: ``Did this object arrive or change state?" 
    \item If an object always exists $\rightarrow$ Output: ``None" (or the Person).
    \item If a person walks in and then pulls out Object X $\rightarrow$ Output: ``Object X".
    \item If the camera cuts/pans to reveal Object X $\rightarrow$ Output: ``Object X".
    \item If the video contains only static background or nothing interesting $\rightarrow$ Output: ``None".
\end{itemize}

\medskip
\# \textbf{OUTPUT FORMAT} \\
Strictly use the following json format.

\vspace{0.5em}
\texttt{```json} \\
\texttt{\{} \\
\texttt{\ \ \ "Reasoning":"[Explain why the object is effective]",} \\
\texttt{\ \ \ "Object":"[Name of the target]"} \\
\texttt{\}} \\
\texttt{```}

\end{tcolorbox}

\begin{tcolorbox}[
    title=\textbf{System Prompt for Action Annotation},
    colback=gray!5,
    colframe=gray!60!black,
    breakable,
    boxrule=0.8pt,
    fonttitle=\bfseries,
    left=1em, right=1em, top=1em, bottom=1em
]
\raggedright

\# \textbf{Role} \\
You are an expert in Video Understanding and Event Logic Reasoning. Your task is to generate high-quality, complex user instructions based on the provided video scenario descriptions.

\medskip
\# \textbf{Task Definition}
\begin{enumerate}[label=\arabic*., leftmargin=2em, nosep]
    \item Analyze the provided video scenario or event description.
    \item Identify specific Complex Events. A complex event must involve one or more of the following:
    \begin{itemize}[label=-, leftmargin=1.5em, nosep]
        \item Temporal Logic: Sequence of actions (e.g., before, after, while, afterwards).
        \item Causality: One action causing another.
        \item Multi-object Interaction: Interactions between different objects or people.
        \item Attribute Identification: Specific visual details (clothing, colors, environment).
        \item Audio/Dialogue Context: Specific conversation topics or sound cues associated with an action.
    \end{itemize}
    \item Formulate a natural language User Instruction that a user would ask to find or describe this event.
\end{enumerate}

\medskip
\# \textbf{Instruction Style \& Patterns} \\
Generate natural, varied user queries. Do not stick to a single formula. Use the following patterns as references:
\begin{itemize}[label=-, leftmargin=1.5em, nosep]
    \item Temporal Sequences: ``Show the part where two men in matching blue sweatshirts eat McDonalds and walk together afterwards."
    \item Visual + Dialogue Cues: ``Identify the moment in the video when a woman is browsing a makeup selection... indicated by her comments about items being hers and mentioning moisturizer."
    \item Simultaneous Actions \& Topics: ``Alert me when two people are talking about payment methods like cash and credit while playing a fighting game."
    \item Specific Attributes \& Actions: ``Notify me when the man in a gray top is walking around a garden in the video." or ``Signal me when the video shows the woman in a black coat grooming a horse."
\end{itemize}

\medskip
\# \textbf{Guidelines}
\begin{itemize}[label=-, leftmargin=1.5em, nosep]
    \item The output user instruction must be clear and concise, summarizing the core event of video in one sentence.
    \item Use varied imperative verbs (e.g., Notify me, Alert me, Signal me, Identify, Show, $\dots$)
    \item Focus on the ACTION and INTERACTION (e.g., ``cutting the apple", ``holding the device"), rather than just static attributes.
\end{itemize}

\medskip
\# \textbf{Constraints}
\begin{itemize}[label=-, leftmargin=1.5em, nosep]
    \item Do NOT use JSON, XML, or Markdown code blocks for the output.
    \item Do NOT include timestamps in the output.
    \item Do NOT use bolding symbols (asterisks) in the output.
    \item The Object in the output format represents the User Instruction.
    \item The Reasoning must explain the logical complexity (causality, temporal order, audio-visual grounding, etc.).
\end{itemize}

\medskip
\# \textbf{OUTPUT FORMAT} \\
Strictly use the following json format.

\vspace{0.5em}
\texttt{```json} \\
\texttt{\{} \\
\texttt{\ \ \ "Reasoning":"Explain why this instruction effectively targets the event logic",} \\
\texttt{\ \ \ "Object":"Concise User Instruction for the event"} \\
\texttt{\}} \\
\texttt{```}

\end{tcolorbox}

\subsection{Time Annotation}
\label{Time_Annotation}
\begin{tcolorbox}[
    title=\textbf{System Prompt for High-Precision Time Annotation (Object)},
    colback=gray!5,
    colframe=gray!60!black,
    breakable,
    boxrule=0.8pt,
    fonttitle=\bfseries,
    left=1em, right=1em, top=1em, bottom=1em
]
\raggedright

You are a High-Precision Visual Verification System for a Proactive Alert Dataset. \\
Your specific task is to examine a single video frame and verify the presence of a specific \textbf{Target Object}.

\medskip
\# \textbf{INPUT CONTEXT}
\begin{enumerate}[label=\arabic*., leftmargin=2em, nosep]
    \item Target Object: \{target\_object\}
    \item Visual Input: You will receive a single image frame extracted from a video stream.
\end{enumerate}

\medskip
\# \textbf{YOUR MISSION} \\
Determine if the Target Object (``\{target\_object\}") is clearly visible in the current frame to warrant a user alert.

\medskip
\# \textbf{JUDGMENT CRITERIA (Strict Rules)} \\
To output a positive alert, the object must meet the following conditions:

\smallskip
1. \textbf{Visibility:} The object must be visually identifiable. Do not guess based on context.
\begin{itemize}[label=*, leftmargin=1.5em, nosep]
    \item Acceptable: The object is partially occluded (e.g., a hand holding a knife handle) but strictly recognizable.
    \item Reject: The object is too blurry, too small (pixel-level noise), or completely hidden.
\end{itemize}

\smallskip
2. \textbf{Relevance:}
\begin{itemize}[label=*, leftmargin=1.5em, nosep]
    \item The object should be the actual physical object, not a drawing or icon.
    \item If the Target is ``knife", do not trigger for ``spoon" or ``fork".
\end{itemize}

\smallskip
3. \textbf{Presence:}
\begin{itemize}[label=*, leftmargin=1.5em, nosep]
    \item If the object has left the frame or hasn't entered yet, it is \textless Silence\textgreater.
\end{itemize}

\medskip
\# \textbf{OUTPUT FORMAT} \\
You must strictly output \textbf{ONLY} one of the following two text strings. Do not add punctuation, explanations, or other text.

\medskip
\textbf{Option 1 (Positive):} \\
\textless Attention\textgreater

\medskip
\textbf{Option 2 (Negative):} \\
\textless Silence\textgreater

\end{tcolorbox}

\begin{tcolorbox}[
    title=\textbf{System Prompt for High-Precision Time Annotation (Action)},
    colback=gray!5,
    colframe=gray!60!black,
    breakable,
    boxrule=0.8pt,
    fonttitle=\bfseries,
    left=1em, right=1em, top=1em, bottom=1em
]
\raggedright

You are a High-Precision Visual Verification System for a Proactive Alert Dataset. \\
Your specific task is to examine a single video frame and verify the presence of a specific \textbf{Target Object}.

\medskip
\# \textbf{INPUT CONTEXT}
\begin{enumerate}[label=\arabic*., leftmargin=2em, nosep]
    \item Target Object: \{target\_object\}
    \item Visual Input: You will receive a single image frame extracted from a video stream.
\end{enumerate}

\medskip
\# \textbf{YOUR MISSION} \\
Determine if the Target Object (``\{target\_object\}") is clearly visible in the current frame to warrant a user alert.

\medskip
\# \textbf{JUDGMENT CRITERIA (Strict Rules)} \\
To output a positive alert, the object must meet the following conditions:

\smallskip
1. \textbf{Visibility:} The object must be visually identifiable. Do not guess based on context.
\begin{itemize}[label=*, leftmargin=1.5em, nosep]
    \item Acceptable: The object is partially occluded (e.g., a hand holding a knife handle) but strictly recognizable.
    \item Reject: The object is too blurry, too small (pixel-level noise), or completely hidden.
\end{itemize}

\smallskip
2. \textbf{Relevance:}
\begin{itemize}[label=*, leftmargin=1.5em, nosep]
    \item The object should be the actual physical object, not a drawing or icon.
    \item If the Target is ``knife", do not trigger for ``spoon" or ``fork".
\end{itemize}

\smallskip
3. \textbf{Presence:}
\begin{itemize}[label=*, leftmargin=1.5em, nosep]
    \item If the object has left the frame or hasn't entered yet, it is \textless Silence\textgreater.
\end{itemize}

\medskip
\# \textbf{OUTPUT FORMAT} \\
You must strictly output \textbf{ONLY} one of the following two text strings. Do not add punctuation, explanations, or other text.

\medskip
\textbf{Option 1 (Positive):} \\
\textless Attention\textgreater

\medskip
\textbf{Option 2 (Negative):} \\
\textless Silence\textgreater

\end{tcolorbox}

\begin{tcolorbox}[
    title=\textbf{System Prompt for Time Filter (Object)},
    colback=gray!5,
    colframe=gray!60!black,
    breakable,
    boxrule=0.8pt,
    fonttitle=\bfseries,
    left=1em, right=1em, top=1em, bottom=1em
]
\raggedright

You are a Logic Consistency Validator for a Security Alert System. \\
Your task is to verify if the System Output matches the Visual Reality.

\medskip
\# \textbf{DEFINITIONS}
\begin{enumerate}[label=\arabic*., leftmargin=2em, nosep]
    \item Target Object: ``\{target\_object\}"
    \item Current Label: ``\{current\_label\}"
    \begin{itemize}[label=-, leftmargin=1.5em, nosep]
        \item ``\textless Attention\textgreater" means: Object is DETECTED.
        \item ``\textless Silence\textgreater" means: Object is ABSENT.
    \end{itemize}
\end{enumerate}

\medskip
\# \textbf{VALIDATION LOGIC}
\begin{itemize}[label=-, leftmargin=1.5em, nosep]
    \item IF Object is Visible $\rightarrow$ Label MUST be ``\textless Attention\textgreater".
    \item IF Object is NOT Visible $\rightarrow$ Label MUST be ``\textless Silence\textgreater".
\end{itemize}

\medskip
\# \textbf{OUTPUT FORMAT} \\
Analysis: [Briefly explain what you see and if it matches the label] \\
Final Decision: [GOOD or BAD]
\end{tcolorbox}

\begin{tcolorbox}[
    title=\textbf{System Prompt for Time Filter (Action)},
    colback=gray!5,
    colframe=gray!60!black,
    breakable,
    boxrule=0.8pt,
    fonttitle=\bfseries,
    left=1em, right=1em, top=1em, bottom=1em
]
\raggedright

You are a Logic Consistency Validator for a Complex Event Alert System. \\
Your task is to verify if the \textbf{System Output} generally matches the \textbf{Visual Reality}.

\medskip
\# \textbf{INPUT CONTEXT}
\begin{enumerate}[label=\arabic*., leftmargin=2em, nosep]
    \item User Instruction: ``\{target\_object\}"
    \item Visual Input: Single video frame.
    \item Current Label: ``\{current\_label\}"
\end{enumerate}

\medskip
\# \textbf{VALIDATION LOGIC MATRIX}

\smallskip
\textbf{CASE A: Label is ``\textless Attention\textgreater"}
\begin{itemize}[label=-, leftmargin=1.5em, nosep]
    \item VALID (GOOD):
    \begin{itemize}[label=$\circ$, leftmargin=1.5em, nosep]
        \item The event described is visually relevant in the frame.
        \item Tolerance: It is ACCEPTABLE if the action is just starting (initiation) or just finishing (follow-through).
        \item \textbf{Close-ups/Occlusion:} The frame shows the essential object or interaction, even if the main actor is cropped out. This is a VALID positive.
    \end{itemize}
    \item INVALID (BAD -- Hallucination/Error):
    \begin{itemize}[label=$\circ$, leftmargin=1.5em, nosep]
        \item Total Irrelevance: The event is clearly NOT happening.
        \item Attribute Mismatch: The action is right, but the person/object is wrong.
    \end{itemize}
\end{itemize}

\smallskip
\textbf{CASE B: Label is ``\textless Silence\textgreater"}
\begin{itemize}[label=-, leftmargin=1.5em, nosep]
    \item VALID (GOOD):
    \begin{itemize}[label=$\circ$, leftmargin=1.5em, nosep]
        \item The frame is clearly unrelated to the user instruction.
        \item The subject is absent, or the scene is static/idle long before or after the event.
    \end{itemize}
    \item INVALID (BAD -- Missed Detection):
    \begin{itemize}[label=$\circ$, leftmargin=1.5em, nosep]
        \item The event is clearly and undeniably occurring, but the system output is ``\textless Silence\textgreater".
    \end{itemize}
\end{itemize}

\medskip
\# \textbf{JUDGMENT PRINCIPLE: ``Semantic Relevance"}
\begin{itemize}[label=-, leftmargin=1.5em, nosep]
    \item Do NOT be overly pedantic about exact time of ``start" or ``end".
    \item Rule of Thumb: If a human looking at this frame would say, ``Yes, this is part of that event," then ``\textless Attention\textgreater" is Correct.
    \item Only mark it as BAD if the system is blatantly wrong.
\end{itemize}

\medskip
\# \textbf{OUTPUT FORMAT} \\
Strictly use the following format.

\medskip
Analysis: [Briefly describe if the visual content is semantically relevant to the label.] \\
Final Decision: [GOOD or BAD]

\end{tcolorbox}

\subsection{Domain Category}
\label{sec:domain_category}

\begin{tcolorbox}[
    title=\textbf{System Prompt for Domain Classification},
    colback=gray!5,
    colframe=gray!60!black,
    breakable,
    boxrule=0.8pt,
    fonttitle=\bfseries,
    left=1em, right=1em, top=1em, bottom=1em
]
\raggedright

\# \textbf{Your Role} \\
You are a \textbf{Domain Analyst}. Your expertise lies in identifying the most accurate match for a video caption from a provided domain list and generating a domain-specific background description.

\medskip
\# \textbf{Your Task}
\begin{itemize}[label=-, leftmargin=1.5em, nosep]
    \item Provide a one-sentence summary of the video.
    \item \textbf{Identify} the specific domain from the provided list.
    \item Generate a \textbf{detailed scene background description} (\texttt{domain\_specific\_description}).
\end{itemize}

\medskip
\# \textbf{Guidelines}
\begin{itemize}[label=-, leftmargin=1.5em, nosep]
    \item The output must be objective and analytical.
    \item The domain \textbf{must be strictly} selected from the provided list.
    \item Start the description with: ``This is a [Domain Name] scene, the user..."
\end{itemize}

\medskip
\# \textbf{Expression Style Restrictions}
\begin{itemize}[label=-, leftmargin=1.5em, nosep]
    \item Use specific, colloquial, and factual language.
    \item Do not use metaphors, symbols, or literary expressions.
\end{itemize}

\medskip
\# \textbf{Word Count and Structure (Mandatory)}
\begin{itemize}[label=-, leftmargin=1.5em, nosep]
    \item Total word count must be strictly controlled between \textbf{25--35 words}.
    \item Reduce event details rather than expanding if clarity cannot be achieved within limits.
\end{itemize}

\medskip
\# \textbf{JSON OUTPUT FORMAT} \\
The response must \textbf{only use JSON format}. The JSON must contain the fields \texttt{domain} and \texttt{domain\_specific\_description}.

\vspace{0.5em}
\texttt{\{} \\
\texttt{\ \ \ \ "domain": "[Most appropriate domain]",} \\
\texttt{\ \ \ \ "domain\_specific\_description": "This is a [Domain Name] scene, [Description]"} \\
\texttt{\}}

\medskip
Output JSON immediately.

\end{tcolorbox}

\subsection{Domain Description}
\label{sec:domain_description}

\begin{tcolorbox}[
    title=\textbf{System Prompt for Domain Description},
    colback=gray!5,
    colframe=gray!60!black,
    breakable,
    boxrule=0.8pt,
    fonttitle=\bfseries,
    left=1em, right=1em, top=1em, bottom=1em
]
\raggedright

\# \textbf{Your Role} \\
You are a \textbf{Domain Analyst}, skilled at generating accurate and concise scene backgrounds from a given domain and a short video summary.

\medskip
\# \textbf{Domain} \\
\{domain\}

\medskip
\# \textbf{Your Task}
\begin{itemize}[label=-, leftmargin=1.5em, nosep]
    \item Summarize the video in one sentence.
    \item Based on the domain, generate a \textbf{detailed scene description} (\texttt{domain\_specific\_description}) in English.
\end{itemize}

\medskip
\# \textbf{Guidelines}
\begin{itemize}[label=-, leftmargin=1.5em, nosep]
    \item The output must be objective and analytical.
    \item Focus on the domain name; start the description with: ``This is a [domain] scene, the user..."
    \item Do not use metaphors, symbols, or literary expressions.
\end{itemize}

\medskip
\# \textbf{Word Count and Structure (Mandatory)}
\begin{itemize}[label=-, leftmargin=1.5em, nosep]
    \item Strictly \textbf{25--35 words}.
    \item If it's impossible to express fully within this word count, reduce details rather than expand.
\end{itemize}

\medskip
\# \textbf{JSON OUTPUT FORMAT} \\
Output must be \textbf{strictly JSON only}. The JSON must contain the field \texttt{domain\_specific\_description}.

\vspace{0.5em}
\texttt{\{} \\
\texttt{\ \ \ \ "domain\_specific\_description": "This is a [domain] scene, the user [factual description]"} \\
\texttt{\}}

\medskip
Output JSON immediately.

\end{tcolorbox}

\subsection{User Memory Generation}

\label{sec:memory_generate_and_filter}
\begin{tcolorbox}[
    title=\textbf{System Prompt for User Memory Generation (Navigation)},
    colback=gray!5,
    colframe=gray!60!black,
    breakable,
    boxrule=0.8pt,
    fonttitle=\bfseries,
    left=1em, right=1em, top=1em, bottom=1em
]
\raggedright

\# \textbf{Role and Task} \\
You are a concise robot. Your task is to summarize the blind person's purpose in the video.

\medskip
\# \textbf{Domain Description}
\begin{itemize}[label=-, leftmargin=1.5em, nosep]
    \item \textbf{Domain Category}: The user is blind, summarize their purpose (destination, intention).
    \item \textbf{Generation Notice}: Ensure the generated \texttt{USER\_MEMORY} clearly indicates the user's blindness. Ignore any subtitles in the video.
\end{itemize}

\medskip
\# \textbf{(User Intention) USER\_MEMORY Construction Principles}

\smallskip
1. \textbf{Intention Nature}
\begin{itemize}[label=*, leftmargin=1.5em, nosep]
    \item Must be directly related to the current video (e.g., attending a concert, taking the subway).
    \item Focus on one primary point only, even if multiple scenes appear.
\end{itemize}

\smallskip
2. \textbf{Expression Style Restrictions}
\begin{itemize}[label=*, leftmargin=1.5em, nosep]
    \item Use specific, colloquial, factual language.
    \item Do not use metaphors, symbols, or literary expressions.
\end{itemize}

\smallskip
3. \textbf{Word Count and Structure (Mandatory)}
\begin{itemize}[label=*, leftmargin=1.5em, nosep]
    \item Total word count must be strictly limited to \textbf{25--35 words}.
    \item If clarity cannot be achieved, reduce event details rather than expanding.
\end{itemize}

\smallskip
4. \textbf{Notes}
\begin{itemize}[label=*, leftmargin=1.5em, nosep]
    \item Use the third person (e.g., ``The user is going to a concert").
    \item Do not describe potential future situations; focus on the current purpose.
\end{itemize}

\medskip
\# \textbf{OUTPUT FORMAT} \\
Output only valid JSON with no additional text.

\vspace{0.5em}
\texttt{\{} \\
\texttt{\ \ \ \ "USER\_MEMORY": "The user is blind + normal memory..."} \\
\texttt{\}}

\end{tcolorbox}
\begin{tcolorbox}[
    title=\textbf{System Prompt for User Memory Generation (Cooking)},
    colback=gray!5,
    colframe=gray!60!black,
    breakable,
    boxrule=0.8pt,
    fonttitle=\bfseries,
    left=1em, right=1em, top=1em, bottom=1em
]
\raggedright

\# \textbf{Role and Task} \\
You act as a third-person narrator, describing an individual's experiences. Your task is to generate an individual's experiences related to cooking, combining user information with environmental background.

\medskip
\# \textbf{Input Data Description}
\begin{enumerate}[label=\arabic*., leftmargin=2em, nosep]
    \item User's Information: \$\{user\}
    \item Domain Description: Cooking/kitchen scene
    \item \texttt{USER\_MEMORY} Generation Notes (Notice)
\end{enumerate}

\medskip
\# \textbf{USER\_MEMORY Generation Notes (Notice)}

\smallskip
\textbf{Content Strategy:}
\begin{itemize}[label=-, leftmargin=1.5em, nosep]
    \item Focus on \textbf{one primary point} from the video or user information.
    \item Do not describe current video content; relate it to a \textbf{prior memory} (days, weeks, or months ago).
    \item Avoid experiences that occurred within the last 72 hours.
    \item Ignore all on-screen text and subtitles.
\end{itemize}

\smallskip
\textbf{Expression Style:}
\begin{itemize}[label=-, leftmargin=1.5em, nosep]
    \item Use specific, colloquial, factual language. No metaphors or literary expressions.
    \item Start with ``The user," (e.g., ``The user tried a certain recipe before...").
\end{itemize}

\smallskip
\textbf{Length and Structure:}
\begin{itemize}[label=-, leftmargin=1.5em, nosep]
    \item Strictly limit total word count to \textbf{10--30 words}.
    \item Use concise statements; complex sentences are not required.
\end{itemize}

\medskip
\# \textbf{OUTPUT FORMAT} \\
Output only valid JSON with no additional text.

\vspace{0.5em}
\texttt{\{} \\
\texttt{\ \ \ \ "USER\_MEMORY": "..."} \\
\texttt{\}}

\end{tcolorbox}

\begin{tcolorbox}[
    title=\textbf{System Prompt for User Memory Generation (Shopping)},
    colback=gray!5,
    colframe=gray!60!black,
    breakable,
    boxrule=0.8pt,
    fonttitle=\bfseries,
    left=1em, right=1em, top=1em, bottom=1em
]
\raggedright

\# \textbf{Role and Task} \\
You act as a third-person narrator, describing an individual's experiences. Your task is to generate an individual's experiences related to shopping, combining user information with environmental background.

\medskip
\# \textbf{Input Data Description}
\begin{enumerate}[label=\arabic*., leftmargin=2em, nosep]
    \item User's Information: \$\{user\}
    \item Domain Description: Shopping scene
    \item \texttt{USER\_MEMORY} Generation Notes (Notice)
\end{enumerate}

\medskip
\# \textbf{USER\_MEMORY Generation Notes (Notice)}

\smallskip
\textbf{Content Strategy:}
\begin{itemize}[label=-, leftmargin=1.5em, nosep]
    \item Focus on \textbf{one primary point} from the video or user information.
    \item Do not describe current video content; relate it to a \textbf{prior memory} (e.g., recently bought a car and needs accessories).
    \item Avoid experiences that occurred within the last 72 hours.
    \item Do \textbf{NOT} describe what the user learned or how they act in the current scene.
    \item Ignore all on-screen text and subtitles.
\end{itemize}

\smallskip
\textbf{Expression Style:}
\begin{itemize}[label=-, leftmargin=1.5em, nosep]
    \item Use specific, colloquial, factual language. No metaphors or literary expressions.
    \item Start with ``The user," (e.g., ``The user recently bought a new car...").
\end{itemize}

\smallskip
\textbf{Length and Structure:}
\begin{itemize}[label=-, leftmargin=1.5em, nosep]
    \item Strictly limit total word count to \textbf{10--30 words}.
    \item Use concise statements or phrases; complex sentences are not required.
\end{itemize}

\medskip
\# \textbf{OUTPUT FORMAT} \\
Output only valid JSON with no additional text.

\vspace{0.5em}
\texttt{\{} \\
\texttt{\ \ \ \ "USER\_MEMORY": "..."} \\
\texttt{\}}

\end{tcolorbox}

\begin{tcolorbox}[
    title=\textbf{System Prompt for User Memory Generation (Entertainment)},
    colback=gray!5,
    colframe=gray!60!black,
    breakable,
    boxrule=0.8pt,
    fonttitle=\bfseries,
    left=1em, right=1em, top=1em, bottom=1em
]
\raggedright

\# \textbf{Role and Task} \\
You act as a third-person narrator, describing an individual's experiences. Your task is to generate an individual's experiences related to entertainment/games, combining user information with environmental background.

\medskip
\# \textbf{Input Data Description}
\begin{enumerate}[label=\arabic*., leftmargin=2em, nosep]
    \item User's Information: \$\{user\}
    \item Domain Description: Entertainment/Game scene
    \item \texttt{USER\_MEMORY} Generation Notes (Notice)
\end{enumerate}

\medskip
\# \textbf{USER\_MEMORY Generation Notes (Notice)}

\smallskip
\textbf{Content Strategy:}
\begin{itemize}[label=-, leftmargin=1.5em, nosep]
    \item Focus on \textbf{one primary point} from the video or user information.
    \item Do not describe current video content; relate it to a \textbf{prior memory} (e.g., learned table tennis when young).
    \item Avoid experiences that occurred within the last 72 hours.
    \item Do \textbf{NOT} describe what the user learned/realized or how they act in the current scene.
    \item Ignore all on-screen text and subtitles.
\end{itemize}

\smallskip
\textbf{Expression Style:}
\begin{itemize}[label=-, leftmargin=1.5em, nosep]
    \item Use specific, colloquial, factual language. No metaphors or literary expressions.
    \item Start with ``The user," (e.g., ``The user likes to play strategy games...").
\end{itemize}

\smallskip
\textbf{Length and Structure:}
\begin{itemize}[label=-, leftmargin=1.5em, nosep]
    \item Strictly limit total word count to \textbf{10--30 words}.
    \item Use concise statements or phrases; complex sentences are not required.
\end{itemize}

\medskip
\# \textbf{OUTPUT FORMAT} \\
Output only valid JSON with no additional text.

\vspace{0.5em}
\texttt{\{} \\
\texttt{\ \ \ \ "USER\_MEMORY": "..."} \\
\texttt{\}}

\end{tcolorbox}

\begin{tcolorbox}[
    title=\textbf{System Prompt for User Memory Generation (Working)},
    colback=gray!5,
    colframe=gray!60!black,
    breakable,
    boxrule=0.8pt,
    fonttitle=\bfseries,
    left=1em, right=1em, top=1em, bottom=1em
]
\raggedright

\# \textbf{Role and Task} \\
You act as a third-person narrator, describing an individual's experiences. Your task is to generate an individual's experiences related to working, combining user information with environmental background.

\medskip
\# \textbf{Input Data Description}
\begin{enumerate}[label=\arabic*., leftmargin=2em, nosep]
    \item User's Information: \$\{user\}
    \item Domain Description: Working/manual/task scene
    \item \texttt{USER\_MEMORY} Generation Notes (Notice)
\end{enumerate}

\medskip
\# \textbf{USER\_MEMORY Generation Notes (Notice)}

\smallskip
\textbf{Content Strategy:}
\begin{itemize}[label=-, leftmargin=1.5em, nosep]
    \item Focus on \textbf{one primary point} from the video or user information.
    \item Do not describe current video content; relate it to a \textbf{prior memory} (e.g., learning a new programming language recently).
    \item Avoid experiences that occurred within the last 72 hours.
    \item Do \textbf{NOT} describe what the user learned/realized or how they act in the current scene.
    \item Ignore all on-screen text and subtitles.
\end{itemize}

\smallskip
\textbf{Expression Style:}
\begin{itemize}[label=-, leftmargin=1.5em, nosep]
    \item Use specific, colloquial, factual language. No metaphors or literary expressions.
    \item Start with ``The user," (e.g., ``The user likes to assemble furniture themselves...").
\end{itemize}

\smallskip
\textbf{Length and Structure:}
\begin{itemize}[label=-, leftmargin=1.5em, nosep]
    \item Strictly limit total word count to \textbf{10--30 words}.
    \item Use concise statements or phrases; complex sentences are not required.
\end{itemize}

\medskip
\# \textbf{OUTPUT FORMAT} \\
Output only valid JSON with no additional text.

\vspace{0.5em}
\texttt{\{} \\
\texttt{\ \ \ \ "USER\_MEMORY": "..."} \\
\texttt{\}}

\end{tcolorbox}

\begin{tcolorbox}[
    title=\textbf{System Prompt for User Memory Generation (Dailylife)},
    colback=gray!5,
    colframe=gray!60!black,
    breakable,
    boxrule=0.8pt,
    fonttitle=\bfseries,
    left=1em, right=1em, top=1em, bottom=1em
]
\raggedright

\# \textbf{Role and Task} \\
You act as a third-person narrator, describing an individual's experiences. Your task is to generate an individual's experiences related to daily life, combining user information with environmental background.

\medskip
\# \textbf{Input Data Description}
\begin{enumerate}[label=\arabic*., leftmargin=2em, nosep]
    \item User's Information: \$\{user\}
    \item Domain Description: Daily life (casual chat/emotional care)
    \item \texttt{USER\_MEMORY} Generation Notes (Notice)
\end{enumerate}

\medskip
\# \textbf{USER\_MEMORY Generation Notes (Notice)}

\smallskip
\textbf{Content Strategy:}
\begin{itemize}[label=-, leftmargin=1.5em, nosep]
    \item Focus on \textbf{one primary point} from the video or user information.
    \item Do not describe current video content; relate it to a \textbf{prior memory} (e.g., went to see a doctor for back pain a few days ago).
    \item Avoid experiences that occurred within the last 72 hours.
    \item Do \textbf{NOT} describe what the user learned/realized or how they act in the current scene.
    \item Ignore all on-screen text and subtitles.
\end{itemize}

\smallskip
\textbf{Expression Style:}
\begin{itemize}[label=-, leftmargin=1.5em, nosep]
    \item Use specific, colloquial, factual language. No metaphors or literary expressions.
    \item Start with ``The user," (e.g., ``The user wants to find a girlfriend").
\end{itemize}

\smallskip
\textbf{Length and Structure:}
\begin{itemize}[label=-, leftmargin=1.5em, nosep]
    \item Strictly limit total word count to \textbf{10--30 words}.
    \item Use concise statements or phrases; complex sentences are not required.
\end{itemize}

\medskip
\# \textbf{OUTPUT FORMAT} \\
Output only valid JSON with no additional text.

\vspace{0.5em}
\texttt{\{} \\
\texttt{\ \ \ \ "USER\_MEMORY": "..."} \\
\texttt{\}}

\end{tcolorbox}

\begin{tcolorbox}[
    title=\textbf{System Prompt for User Memory Generation (Travel)},
    colback=gray!5,
    colframe=gray!60!black,
    breakable,
    boxrule=0.8pt,
    fonttitle=\bfseries,
    left=1em, right=1em, top=1em, bottom=1em
]
\raggedright

\# \textbf{Role and Task} \\
You act as a third-person narrator, describing an individual's experiences. Your task is to generate an individual's experiences related to travel, combining user information with environmental background.

\medskip
\# \textbf{Input Data Description}
\begin{enumerate}[label=\arabic*., leftmargin=2em, nosep]
    \item User's Information: \$\{user\}
    \item Domain Description: Travel/life (casual chat/emotional care)
    \item \texttt{USER\_MEMORY} Generation Notes (Notice)
\end{enumerate}

\medskip
\# \textbf{USER\_MEMORY Generation Notes (Notice)}

\smallskip
\textbf{Content Strategy:}
\begin{itemize}[label=-, leftmargin=1.5em, nosep]
    \item Focus on \textbf{one primary point} from the video or user information.
    \item Do not describe current video content; relate it to a \textbf{prior memory} (e.g., preference for local food when traveling).
    \item Avoid experiences that occurred within the last 72 hours.
    \item Do \textbf{NOT} describe what the user learned/realized or how they act in the current scene.
    \item Ignore all on-screen text and subtitles.
\end{itemize}

\smallskip
\textbf{Expression Style:}
\begin{itemize}[label=-, leftmargin=1.5em, nosep]
    \item Use specific, colloquial, factual language. No metaphors or literary expressions.
    \item Start with ``The user," (e.g., ``The user likes to bring back many souvenirs...").
\end{itemize}

\smallskip
\textbf{Length and Structure:}
\begin{itemize}[label=-, leftmargin=1.5em, nosep]
    \item Strictly limit total word count to \textbf{10--30 words}.
    \item Use concise statements or phrases; complex sentences are not required.
\end{itemize}

\medskip
\# \textbf{OUTPUT FORMAT} \\
Output only valid JSON with no additional text.

\vspace{0.5em}
\texttt{\{} \\
\texttt{\ \ \ \ "USER\_MEMORY": "..."} \\
\texttt{\}}

\end{tcolorbox}

\begin{tcolorbox}[
    title=\textbf{System Prompt for User Memory Generation (Art)},
    colback=gray!5,
    colframe=gray!60!black,
    breakable,
    boxrule=0.8pt,
    fonttitle=\bfseries,
    left=1em, right=1em, top=1em, bottom=1em
]
\raggedright

\# \textbf{Role and Task} \\
You act as a third-person narrator, describing an individual's experiences. Your task is to generate an individual's experiences related to painting/artistic creation, combining user information with environmental background.

\medskip
\# \textbf{Input Data Description}
\begin{enumerate}[label=\arabic*., leftmargin=2em, nosep]
    \item User's Information: \$\{user\}
    \item Domain Description: Painting/artistic creation scene
    \item \texttt{USER\_MEMORY} Generation Notes (Notice)
\end{enumerate}

\medskip
\# \textbf{USER\_MEMORY Generation Notes (Notice)}

\smallskip
\textbf{Content Strategy:}
\begin{itemize}[label=-, leftmargin=1.5em, nosep]
    \item Focus on \textbf{one primary point} from the video or user information.
    \item Do not describe current video content; relate it to a \textbf{prior memory} (e.g., learned to paint when young).
    \item Avoid experiences that occurred within the last 72 hours.
    \item Do \textbf{NOT} describe what the user learned/realized or how they act in the current scene.
    \item Ignore all on-screen text and subtitles.
\end{itemize}

\smallskip
\textbf{Expression Style:}
\begin{itemize}[label=-, leftmargin=1.5em, nosep]
    \item Use specific, colloquial, factual language. No metaphors or literary expressions.
    \item Start with ``The user," (e.g., ``The user likes to create landscapes with oil paints...").
\end{itemize}

\smallskip
\textbf{Length and Structure:}
\begin{itemize}[label=-, leftmargin=1.5em, nosep]
    \item Strictly limit total word count to \textbf{10--30 words}.
    \item Use concise statements or phrases; complex sentences are not required.
\end{itemize}

\medskip
\# \textbf{OUTPUT FORMAT} \\
Output only valid JSON with no additional text.

\vspace{0.5em}
\texttt{\{} \\
\texttt{\ \ \ \ "USER\_MEMORY": "..."} \\
\texttt{\}}

\end{tcolorbox}

\begin{tcolorbox}[
    title=\textbf{System Prompt for User Memory Generation (Sports)},
    colback=gray!5,
    colframe=gray!60!black,
    breakable,
    boxrule=0.8pt,
    fonttitle=\bfseries,
    left=1em, right=1em, top=1em, bottom=1em
]
\raggedright

\# \textbf{Role and Task} \\
You act as a third-person narrator, describing an individual's experiences. Your task is to generate an individual's experiences related to sports, combining user information with environmental background.

\medskip
\# \textbf{Input Data Description}
\begin{enumerate}[label=\arabic*., leftmargin=2em, nosep]
    \item User's Information: \$\{user\}
    \item Domain Description: Sports scene
    \item \texttt{USER\_MEMORY} Generation Notes (Notice)
\end{enumerate}

\medskip
\# \textbf{USER\_MEMORY Generation Notes (Notice)}

\smallskip
\textbf{Content Strategy:}
\begin{itemize}[label=-, leftmargin=1.5em, nosep]
    \item Focus on \textbf{one primary point} from the video or user information.
    \item Do not describe current video content; relate it to a \textbf{prior memory} (e.g., learned table tennis when young).
    \item Avoid experiences that occurred within the last 72 hours.
    \item Do \textbf{NOT} describe what the user learned/realized or how they act in the current scene.
    \item Ignore all on-screen text and subtitles.
\end{itemize}

\smallskip
\textbf{Expression Style:}
\begin{itemize}[label=-, leftmargin=1.5em, nosep]
    \item Use specific, colloquial, factual language. No metaphors or literary expressions.
    \item Start with ``The user," (e.g., ``The user likes to play badminton...").
\end{itemize}

\smallskip
\textbf{Length and Structure:}
\begin{itemize}[label=-, leftmargin=1.5em, nosep]
    \item Strictly limit total word count to \textbf{10--30 words}.
    \item Use concise statements or phrases; complex sentences are not required.
\end{itemize}

\medskip
\# \textbf{OUTPUT FORMAT} \\
Output only valid JSON with no additional text.

\vspace{0.5em}
\texttt{\{} \\
\texttt{\ \ \ \ "USER\_MEMORY": "..."} \\
\texttt{\}}

\end{tcolorbox}

\begin{tcolorbox}[
    title=\textbf{System Prompt for User Memory Generation (Driving)},
    colback=gray!5,
    colframe=gray!60!black,
    breakable,
    boxrule=0.8pt,
    fonttitle=\bfseries,
    left=1em, right=1em, top=1em, bottom=1em
]
\raggedright

\# \textbf{Role and Task} \\
You act as a third-person narrator, describing an individual's experiences. Your task is to generate an individual's experiences related to driving, combining user information with environmental background.

\medskip
\# \textbf{Input Data Description}
\begin{enumerate}[label=\arabic*., leftmargin=2em, nosep]
    \item User's Information: \$\{user\}
    \item Domain Description: Driving scene
    \item \texttt{USER\_MEMORY} Generation Notes (Notice)
\end{enumerate}

\medskip
\# \textbf{USER\_MEMORY Generation Notes (Notice)}

\smallskip
\textbf{Content Strategy:}
\begin{itemize}[label=-, leftmargin=1.5em, nosep]
    \item Focus on \textbf{one primary point} from the video or user information.
    \item Do not describe current video content; relate it to a \textbf{prior memory} (e.g., having had a car accident before).
    \item Avoid experiences that occurred within the last 72 hours.
    \item Do \textbf{NOT} describe what the user learned/realized or how they act in the current scene.
    \item Ignore all on-screen text and subtitles.
\end{itemize}

\smallskip
\textbf{Expression Style:}
\begin{itemize}[label=-, leftmargin=1.5em, nosep]
    \item Use specific, colloquial, factual language. No metaphors or literary expressions.
    \item Start with ``The user," (e.g., ``The user likes to drive carefully...").
\end{itemize}

\smallskip
\textbf{Length and Structure:}
\begin{itemize}[label=-, leftmargin=1.5em, nosep]
    \item Strictly limit total word count to \textbf{10--30 words}.
    \item Use concise statements or phrases; complex sentences are not required.
\end{itemize}

\medskip
\# \textbf{OUTPUT FORMAT} \\
Output only valid JSON with no additional text.

\vspace{0.5em}
\texttt{\{} \\
\texttt{\ \ \ \ "USER\_MEMORY": "..."} \\
\texttt{\}}

\end{tcolorbox}

\subsection{User Memory Filter}
\label{sec:memory_filter}
\begin{tcolorbox}[
    title=\textbf{System Prompt for Memory Filter (Vision)},
    colback=gray!5,
    colframe=gray!60!black,
    breakable,
    boxrule=0.8pt,
    fonttitle=\bfseries,
    left=1em, right=1em, top=1em, bottom=1em
]
\raggedright

You are a User Memory Quality Auditor. Your task is to determine whether a given User Memory can be triggered by the current video content.

\medskip
\# \textbf{Visual Trigger Audit}
\begin{itemize}[label=-, leftmargin=1.5em, nosep]
    \item Determine if the memory can be directly triggered by objects, actions, or scenes visible in the video.
    \item Any memory that fails to establish a reasonable visual association is considered unqualified.
\end{itemize}

\medskip
\# \textbf{Output Requirements} \\
Output \textbf{ONLY} a valid JSON object without any additional text. The structure must be as follows:

\vspace{0.5em}
\texttt{\{} \\
\texttt{\ \ \ \ "Reasoning": \{} \\
\texttt{\ \ \ \ \ \ \ \ "Video Relevance": "<Short English reasoning>"} \\
\texttt{\ \ \ \ \},} \\
\texttt{\ \ \ \ "Score": \{} \\
\texttt{\ \ \ \ \ \ \ \ "Video Relevance": 1|3|5} \\
\texttt{\ \ \ \ \}} \\
\texttt{\}}

\medskip
\# \textbf{Scoring Rules}
\begin{itemize}[label=-, leftmargin=1.5em, nosep]
    \item \textbf{Video Relevance}: Evaluated only when Memory Logical Consistency (from text audit) is 5; otherwise, it must be 1.
    \begin{itemize}[label=$\circ$, leftmargin=1.5em, nosep]
        \item 1 = No trigger: No visual connection found.
        \item 3 = Weak trigger: Indirect or subtle visual connection.
        \item 5 = Strong trigger: Direct and clear visual connection.
    \end{itemize}
\end{itemize}

\medskip
Output JSON immediately.

\end{tcolorbox}

\begin{tcolorbox}[
    title=\textbf{System Prompt for Memory Filter (Text)},
    colback=gray!5,
    colframe=gray!60!black,
    breakable,
    boxrule=0.8pt,
    fonttitle=\bfseries,
    left=1em, right=1em, top=1em, bottom=1em
]
\raggedright

\medskip
You are a User Memory Quality Auditor. Your task is to determine whether a given User Memory is a qualified and usable memory text.

\medskip
\# \textbf{Audit Objective} \\
Evaluate the credibility and everyday plausibility of the memory text itself.

\medskip
\# \textbf{Audit Criteria} \\
Determine if the memory is qualified based on the following:
\begin{itemize}[label=-, leftmargin=1.5em, nosep]
    \item It must describe a specific event or experience that conforms to common sense and daily life.
    \item It must describe past memories or user preferences/habits.
    \item It must \textbf{NOT} describe what the user is currently doing (ongoing actions).
    \item The memory must be in English; no Chinese characters allowed.
\end{itemize}

\medskip
\# \textbf{Output Requirements} \\
Output \textbf{ONLY} a valid JSON object without any additional text. The structure must be as follows:

\vspace{0.5em}
\texttt{\{} \\
\texttt{\ \ \ \ "Reasoning": \{} \\
\texttt{\ \ \ \ \ \ \ \ "Memory Logical Consistency": "<Short English reasoning>"} \\
\texttt{\ \ \ \ \},} \\
\texttt{\ \ \ \ "Score": \{} \\
\texttt{\ \ \ \ \ \ \ \ "Memory Consistency": 1|3|5} \\
\texttt{\ \ \ \ \}} \\
\texttt{\}}

\medskip
\# \textbf{Scoring Rules}
\begin{itemize}[label=-, leftmargin=1.5em, nosep]
    \item 1 = Unqualified: The memory is implausible or logically flawed.
    \item 3 = Acceptable: The memory is plausible but describes an uncommon or rare situation.
    \item 5 = Qualified: The memory is realistic and common in daily life.
\end{itemize}

\medskip
Output JSON immediately.

\medskip
\end{tcolorbox}

\subsection{Response Generation}
\label{sec:response_generate}

\begin{tcolorbox}[
    title=\textbf{System Prompt for Intent Response Generation},
    colback=gray!5,
    colframe=gray!60!black,
    breakable,
    boxrule=0.8pt,
    fonttitle=\bfseries,
    left=1em, right=1em, top=1em, bottom=1em
]
\raggedright

\# \textbf{Your Role} \\
You are an AI assistant. Respond to the user only when necessary.

\medskip
\# \textbf{Core Judgment Perspective}
\begin{itemize}[label=-, leftmargin=1.5em, nosep]
    \item User's memory (\texttt{user\_memory})
    \item Current scene (\texttt{domain\_background})
\end{itemize}

\medskip
\# \textbf{Reference Materials}
\begin{enumerate}[label=\arabic*., leftmargin=2em, nosep]
    \item \texttt{user\_memory}: \$\{user\_memory\}
    \item \texttt{domain\_background}: \$\{domain\_background\}
\end{enumerate}

\medskip
\# \textbf{Response Timing and Response Content}

\smallskip
\textbf{Dimension 1: Response Timing Suitability (1 / 3 / 5)}
\begin{itemize}[label=*, leftmargin=1.5em, nosep]
    \item \textbf{5 points}: Current frame is highly relevant to \texttt{user\_memory}; ideal time to respond.
    \item \textbf{3 points}: Some connection exists, but to avoid disturbance, output \texttt{<Silence>}.
    \item \textbf{1 point}: Completely irrelevant to \texttt{user\_memory}, output \texttt{<Silence>}.
\end{itemize}

\smallskip
\textbf{Dimension 2: Response Content Generation}
\begin{itemize}[label=*, leftmargin=1.5em, nosep]
    \item Generate response only for 5 points; otherwise, generate \texttt{<Silence>}.
    \item Concise response (1--2 sentences) using ``you" to address the user.
\end{itemize}

\medskip
\# \textbf{Notes}
\begin{itemize}[label=-, leftmargin=1.5em, nosep]
    \item Do not describe user actions; provide content directly.
    \item Must be in English.
\end{itemize}

\medskip
\# \textbf{FINAL OUTPUT FORMAT} \\
Output a valid JSON object and nothing else. Strictly follow these fields:

\vspace{0.5em}
\texttt{\{} \\
\texttt{\ \ \ \ "justification": "Analysis reasoning",} \\
\texttt{\ \ \ \ "scores": <1-5>,} \\
\texttt{\ \ \ \ "response\_content": "Reply content or <Silence>"} \\
\texttt{\}}

\medskip
Output JSON immediately.

\end{tcolorbox}

\begin{tcolorbox}[
    title=\textbf{System Prompt for Intent Response Generation (Navigation)},
    colback=gray!5,
    colframe=gray!60!black,
    breakable,
    boxrule=0.8pt,
    fonttitle=\bfseries,
    left=1em, right=1em, top=1em, bottom=1em
]
\raggedright

\# \textbf{Your Role} \\
You are a proactive navigation AI assistant for a blind user. Your primary mission is to ensure safety by identifying potential hazards and guiding the user along clear paths.

\medskip
\# \textbf{Reference Materials}
\begin{enumerate}[label=\arabic*., leftmargin=2em, nosep]
    \item User's current activity: \$\{user\_memory\}
    \item Current scene context: \$\{domain\_background\}
\end{enumerate}

\medskip
\# \textbf{Response Timing and Content Requirements}

\smallskip
\textbf{Dimension 1: Response Timing Suitability (Safety-Critical)}
\begin{itemize}[label=*, leftmargin=1.5em, nosep]
    \item \textbf{5 points}: Obstacle detected within 3 meters; change in terrain (stairs, curbs); tactile paving available; or approaching danger (vehicles/cyclists).
    \item \textbf{1 point}: Path is completely clear for at least 5 meters; ground is flat and safe. Output \texttt{<Silence>}.
\end{itemize}

\smallskip
\textbf{Dimension 2: Response Content Generation}
\begin{itemize}[label=*, leftmargin=1.5em, nosep]
    \item \textbf{Obstacles}: State type, direction (clock position), and approximate distance.
    \item \textbf{Tactile Paving}: Prioritize guiding the user to align with it.
    \item \textbf{Terrain}: Warn about steps, drops, or uneven surfaces.
    \item \textbf{Clarity}: Keep instructions brief and urgent for close obstacles.
\end{itemize}

\medskip
\# \textbf{Notes}
\begin{itemize}[label=-, leftmargin=1.5em, nosep]
    \item Address the user as ``you".
    \item Focus \textbf{ONLY} on the last frame for immediate situational judgment.
    \item Prioritize obstacles intersecting with the user's projected path.
\end{itemize}

\medskip
\# \textbf{FINAL OUTPUT FORMAT} \\
Must output a valid JSON object and only JSON.

\vspace{0.5em}
\texttt{\{} \\
\texttt{\ \ \ \ "justification": "Analysis of obstacles, terrain, and tactile paving.",} \\
\texttt{\ \ \ \ "scores": <1-5>,} \\
\texttt{\ \ \ \ "response\_content": "Directional guidance/warning or <Silence>"} \\
\texttt{\}}

\medskip
Output JSON immediately.

\end{tcolorbox}

\subsection{LLM-as-a-Judge}
\label{llm_judge}
\begin{tcolorbox}[
    title=\textbf{System Prompt for Evaluation of Memory Consistency and Response Quality},
    colback=gray!5,
    colframe=gray!60!black,
    breakable,
    boxrule=0.8pt,
    fonttitle=\bfseries,
    left=1em, right=1em, top=1em, bottom=1em
]
\raggedright

\# \textbf{Your Role and Core Task} \\
You are an expert evaluator specializing in assessing AI model responses. Your core task is to determine whether an AI model's response is appropriate.

\medskip
\# \textbf{Core Evaluation Objectives}
\begin{itemize}[label=-, leftmargin=1.5em, nosep]
    \item Accurately identify whether the AI model's response conflicts with the user's memory.
    \item Score the model's response based on visual information and the provided ground-truth (GT) response.
\end{itemize}

\medskip
\# \textbf{Background}
\begin{itemize}[label=-, leftmargin=1.5em, nosep]
    \item The user input consists of ego-centric video data (Ego-view).
    \item If no response is needed, the AI model outputs \texttt{<Silence>}.
    \item If a response is needed, the AI model generates an appropriate reply.
\end{itemize}

\medskip
\# \textbf{Evaluation Materials}
\begin{enumerate}[label=\arabic*., leftmargin=2em, nosep]
    \item User memory: \{\{User\_Memory\}\}
    \item AI model response: \{\{Model\_Response\}\}
    \item Ground-truth (GT) response: \{\{GT\_Response\}\}
\end{enumerate}

\medskip
\# \textbf{Evaluation Dimensions and Scoring Criteria (1 / 3 / 5)}

\smallskip
\textbf{Dimension 1: Memory Consistency}
\begin{itemize}[label=*, leftmargin=1.5em, nosep]
    \item \textbf{5 points}: No contradiction at all. All memory-related statements are fully consistent.
    \item \textbf{3 points}: Generally reasonable with no obvious or critical conflicts.
    \item \textbf{1 point}: The response clearly contradicts the user's memory.
\end{itemize}

\smallskip
\textbf{Dimension 2: Response Content Quality}
\begin{itemize}[label=*, leftmargin=1.5em, nosep]
    \item \textbf{5 points}: Highly matched in core intent and key information.
    \item \textbf{3 points}: Partially matched; captures some intent but misses or adds a point.
    \item \textbf{1 point}: Completely mismatched. If GT is \texttt{<Silence>}, the model must also be \texttt{<Silence>}.
\end{itemize}

\medskip
\# \textbf{Notes}
\begin{itemize}[label=-, leftmargin=1.5em, nosep]
    \item Only evaluate the \textbf{last frame} of the video for the final state.
    \item Be direct, clear, and concise.
\end{itemize}

\medskip
\# \textbf{OUTPUT FORMAT} \\
Strictly follow the JSON format below.

\vspace{0.5em}
\texttt{\{} \\
\texttt{\ \ \ "justification": \{} \\
\texttt{\ \ \ \ \ \ "Memory Consistency Justification": "<one-sentence justification>",} \\
\texttt{\ \ \ \ \ \ "Response Content Quality Justification": "<one-sentence justification>"} \\
\texttt{\ \ \ \ \},} \\
\texttt{\ \ \ "scores": \{} \\
\texttt{\ \ \ \ \ \ "Memory Consistency": <Score>,} \\
\texttt{\ \ \ \ \ \ "Response Content Quality": <Score>} \\
\texttt{\ \ \ \ \}} \\
\texttt{\}}
\end{tcolorbox}

\subsection{CoT Data Annotation}
\label{appenix_cot}
\begin{tcolorbox}[
    title=\textbf{System Prompt for CoT Annotation (Event-driven)},
    colback=gray!5,
    colframe=gray!60!black,
    breakable,
    boxrule=0.8pt,
    fonttitle=\bfseries,
    left=1em, right=1em, top=1em, bottom=1em
]
\raggedright

You are a visual reasoning expert. Your task is to generate a Chain-of-Thought (CoT) that logically justifies the provided Ground Truth label.

\medskip
\# \textbf{INPUT}
\begin{enumerate}[label=\arabic*., leftmargin=2em, nosep]
    \item User Instruction: The event we are looking for.
    \item Image: The visual evidence.
    \item Ground Truth: ``\textless Attention\textgreater" or ``\textless Silence\textgreater".
\end{enumerate}

\medskip
\# \textbf{REASONING LOGIC (CRITICAL)}

\smallskip
\textbf{Case 1: If Ground Truth is ``\textless Attention\textgreater"}
\begin{itemize}[label=-, leftmargin=1.5em, nosep]
    \item Your goal is to \textbf{find evidence to support} the alert.
    \item \textbf{Handling Partial Visibility (Close-ups):}
    \begin{itemize}[label=$\circ$, leftmargin=1.5em, nosep]
        \item If the instruction mentions ``A man holding a Kindle" but only hands are visible, DO NOT say ``I cannot see the man."
        \item INSTEAD, say: ``I see hands holding a Kindle. Although the actor's body is not fully visible, the interaction confirms the action is in progress."
        \item \textbf{Key Logic}: Visible Core Action $\rightarrow$ Implies Actor Presence $\rightarrow$ Matches Instruction.
    \end{itemize}
\end{itemize}

\smallskip
\textbf{Case 2: If Ground Truth is ``\textless Silence\textgreater"}
\begin{itemize}[label=-, leftmargin=1.5em, nosep]
    \item Describe what is missing (e.g., scene is empty, action is not happening, or object is idle).
\end{itemize}

\medskip
\# \textbf{OUTPUT FORMAT} \\
Output \textbf{ONLY} a valid JSON object:

\vspace{0.5em}
\texttt{\{} \\
\texttt{\ \ \ \ "reasoning": "Concise CoT (Max 30 words).",} \\
\texttt{\ \ \ \ "response": "\textless Attention\textgreater" or "\textless Silence\textgreater"} \\
\texttt{\}}

\medskip
Output JSON immediately.

\end{tcolorbox}

\begin{tcolorbox}[
    title=\textbf{System Prompt for CoT Annotation (Intent-driven)},
    colback=gray!5,
    colframe=gray!60!black,
    breakable,
    boxrule=0.8pt,
    fonttitle=\bfseries,
    left=1em, right=1em, top=1em, bottom=1em
]
\raggedright
You are a visual data annotation assistant. Your goal is to generate a concise, natural Chain-of-Thought (CoT) that simulates the decision-making process based on Visual Content, User Memory, and the Target Response.

\medskip
\# \textbf{Input Context}
\begin{enumerate}[label=\arabic*., leftmargin=2em, nosep]
    \item User Memory: The user's specific habit, history, or preference.
    \item Ground Truth Response: The target output (message or \texttt{<Silence>}).
    \item Relevance Score: A score (0--5) indicating relevance.
    \item Current Image: The visual evidence.
\end{enumerate}

\medskip
\# \textbf{Task} \\
Simulate the inference flow: \textbf{Observe} $\rightarrow$ \textbf{Connect} $\rightarrow$ \textbf{Decide}. \\
Generate a reasoning thought that logically justifies the Ground Truth Response based on the visual evidence and the score.

\medskip
\# \textbf{Reasoning Logic (Choose one based on inputs)}

\smallskip
\textbf{Case 1: Active Response (Score 5 \& GT is text)}
\begin{itemize}[label=-, leftmargin=1.5em, nosep]
    \item Observe $\rightarrow$ Connect $\rightarrow$ Decide: ``I see [action]... which aligns with [Memory]... so I must [provide guidance]."
\end{itemize}

\smallskip
\textbf{Case 2: Strategic Silence (Score 5 \& GT is \texttt{<Silence>})}
\begin{itemize}[label=-, leftmargin=1.5em, nosep]
    \item Logic: The situation matches Memory but is unchanged or repetitive, so I remain silent to avoid being annoying.
\end{itemize}

\smallskip
\textbf{Case 3: No Trigger (Score $<$ 5)}
\begin{itemize}[label=-, leftmargin=1.5em, nosep]
    \item Logic: The visual does not trigger specific concerns related to Memory; no intervention is needed.
\end{itemize}

\medskip
\# \textbf{OUTPUT FORMAT} \\
Output \textbf{ONLY} a valid JSON object:

\vspace{0.5em}
\texttt{\{} \\
\texttt{\ \ \ \ "reasoning": "Your concise CoT reasoning...",} \\
\texttt{\ \ \ \ "response": "The Ground Truth Response"} \\
\texttt{\}}

\medskip
\# \textbf{Guidelines}
\begin{itemize}[label=-, leftmargin=1.5em, nosep]
    \item \textbf{STRICT LIMIT}: Maximum 25 words.
    \item \textbf{Forward-Looking}: Do NOT mention ``Ground Truth" or ``Score" in reasoning.
    \item \textbf{Visual Focus}: Mention specific objects/gestures (e.g., ``holding the breaker").
    \item \textbf{Natural Flow}: Use first-person (``I notice...", ``The user is...").
\end{itemize}

\medskip
Output JSON immediately.

\end{tcolorbox}


\section{User Profile Details}
\label{persona_profile_details}
\subsection{User Occupation List}
\textbf{Healthcare.} Doctor, Surgeon, Internist, Pediatrician, Obstetrician/Gynecologist, Ophthalmologist, Dentist, Nurse, Head Nurse, Midwife, Pharmacist, Veterinarian, Psychiatrist, Psychological Counselor, Physical Therapist, Anesthesiologist, Radiologic Technologist, Medical Laboratory Technician, Dietitian/Nutritionist, Optometrist, Audiologist, Speech Therapist, Caregiver/Nursing Aide, Acupuncturist, Massage Therapist, Chiropractor, Pathologist, Genetic Counselor, Medical Equipment Technician, Public Health Specialist, Epidemiologist, Clinical Research Coordinator, Dental Hygienist, Veterinary Assistant.

\textbf{IT \& Internet.} Software Engineer, Programmer, Front-end Developer, Back-end Developer, Full-stack Developer, Mobile App Developer, Game Developer, DBA, System Administrator, Network Engineer, Information Security Analyst, DevOps Engineer, Cloud Engineer, Data Scientist, Data Analyst, AI Engineer, Machine Learning Engineer, Product Manager, Project Manager, UI Designer, UX Designer, Web Designer, Software QA Engineer, Technical Support Engineer, SEO Specialist, IT Auditor, Embedded Systems Engineer, E-commerce Specialist, Algorithm Engineer, Blockchain Developer, User Researcher, Growth Hacker, Crawler Engineer, New Media Operator.

\textbf{Education \& Research.} Kindergarten Teacher, Elementary School Teacher, Secondary School Teacher, University Professor, Lecturer, Teaching Assistant, Special Education Teacher, Vocational Teacher, Corporate Trainer, Driving Instructor, Fitness Trainer, Yoga Instructor, Music Teacher, Art Teacher, Education Consultant, Principal/Headmaster, Librarian, Archivist, Researcher, Scientist, Historian, Archaeologist, Sociologist, Anthropologist, Astronomer, Physicist, Chemist, Biologist, Geologist, Postdoctoral Researcher, Lab Assistant, Admissions Officer.

\textbf{Business, Finance \& Management.} CEO, COO, CFO, General Manager, Department Manager, Marketing Director, Sales Director, HR Director, Marketing Specialist, Sales Representative, HR Specialist, Recruiter, Administrative Assistant, Office Clerk, Accountant, Auditor, Financial Analyst, Investment Banker, Stock Trader, Fund Manager, Financial Planner, Insurance Agent, Actuary, Real Estate Agent, Purchasing Manager, Logistics Manager, Supply Chain Manager, Operations Manager, Business Consultant, Risk Manager, Tax Advisor, CRM, Brand Manager, Public Relations Specialist, Secretary to the Chairman.

\textbf{Arts, Design \& Media.} Artist, Painter, Sculptor, Photographer, Graphic Designer, Industrial Designer, Fashion Designer, Interior Designer, Landscape Designer, Architectural Designer, Animator, Illustrator, Cartoonist, Writer, Screenwriter, Journalist, Editor, Director, Producer, Actor/Actress, Model, Dancer, Singer, Musician, Composer, Conductor, Host/Presenter, Announcer, Sound Engineer, Lighting Technician, Makeup Artist, Stylist, Curator, Video Editor, Art Director, Copywriter, Colorist, Agent/Manager.

\textbf{Law \& Public Safety.} Lawyer, Judge, Prosecutor, Paralegal, Legal Counsel, Court Clerk, Police Officer, Detective, Firefighter, Correctional Officer, Forensic Scientist/Medical Examiner, Traffic Police, Security Guard, Private Investigator, Customs Officer, Border Patrol Agent, Probation Officer, IP Consultant, Air Marshal, Emergency Dispatcher.

\textbf{Engineering \& Construction.} Architect, Urban Planner, Civil Engineer, Structural Engineer, Mechanical Engineer, Electrical Engineer, Electronics Engineer, Materials Engineer, Chemical Engineer, Environmental Engineer, Aerospace Engineer, Automotive Engineer, Biomedical Engineer, Petroleum Engineer, Mining Engineer, Nuclear Engineer, QC Engineer, Construction Manager, Construction Worker, Bricklayer, Carpenter, Plumber, Electrician, Welder, Painter, Surveyor, Cost Estimator, Crane Operator, Excavator Operator, Draftsperson, Safety Engineer, HVAC Engineer.

\textbf{Transportation \& Logistics.} Pilot, Air Traffic Controller, Flight Attendant, Aircraft Mechanic, Ship Captain, Sailor/Seaman, Navigator, Train Driver, Train Conductor, Subway Operator, Bus Driver, Taxi Driver, Ride-sharing Driver, Truck Driver, Courier/Delivery Driver, Food Delivery Driver, Warehouse Manager, Forklift Operator, Freight Forwarder, Dispatcher, Shipping Operator, Dock Worker, Mover, Ground Crew.

\textbf{Hospitality \& Food Service.} Chef/Cook, Executive Chef, Pastry Chef, Baker, Barista, Bartender, Sommelier, Restaurant Manager, Waiter/Waitress, Busser, Dishwasher, Hotel Manager, Front Desk Clerk, Concierge, Housekeeper, Banquet Manager, Travel Agent, Tour Guide, Ticketing Agent, Croupier, Event Planner.

\textbf{Agriculture, Forestry \& Fishery.} Farmer, Farm Manager, Agricultural Technician, Horticulturist, Florist, Animal Husbandry Specialist, Animal Breeder/Keeper, Fisherman, Aquaculturist, Forestry Worker, Forest Ranger, Soil Scientist, Botanist, Butcher, Winemaker, Agricultural Produce Buyer, Drone Pilot - Agriculture.

\textbf{Manufacturing \& Skilled Trades.} Plant Manager, Production Supervisor, Assembly Line Worker, Machinist, CNC Operator, Mold Designer, Tool and Die Maker, Quality Inspector, Industrial Robotics Engineer, Maintenance Electrician, Mechanic, Auto Mechanic, Sheet Metal Worker, Jeweler/Goldsmith, Watchmaker, Shoemaker, Tailor, Textile Worker, Printer, Bookbinder, Glazier, Locksmith, Cabinetmaker.

\textbf{Sports, Recreation \& Fitness.} Athlete, Coach, Sports Agent, Referee/Umpire, Sports Commentator, Personal Trainer, Sports Physician, Lifeguard, Choreographer, Racing Driver, Scout, Caddie, Diving Instructor, Equestrian, Sports Psychologist.

\textbf{Government, NPO \& Public Service.} Civil Servant, Diplomat, Mayor, Congressperson/Parliamentarian, Mail Carrier, Tax Inspector, Urban Planner, Social Worker, Community Worker, NPO Program Manager, Fundraiser, Environmentalist, Meteorologist, Statistician, Economist, Translator/Interpreter, Museum Director, Fire Inspector, Health Inspector, Ambassador.

\textbf{Personal Services \& Retail.} Barber/Hairstylist, Beautician/Esthetician, Manicurist, Pet Groomer, Dry Cleaner, Shoe Shiner, Mortician/Undertaker, Store Manager, Cashier, Sales Associate, Stocker, Buyer/Merchandiser, Visual Merchandiser, Customer Service Representative, Telemarketer, Pawnbroker, Masseur/Masseuse, Nanny/Babysitter, Housekeeper, Personal Assistant.

\textbf{Other Professions \& Emerging Careers.} Astronaut, Philosopher, Clergy/Priest/Minister, Astrologer, Feng Shui Master, Magician, Clown, Street Performer, Professional Gamer, Vlogger/YouTuber, Live Streamer, Podcaster, Freelancer, Voice Actor, Hotel Tester, Game Companion, Professional Organizer, Drone Operator, 3D Printing Engineer, VR Developer, AR Developer, Sustainability Consultant, Carbon Trader, UX Researcher, Chief Ethics Officer, Gene-editing Specialist, Quantum Computing Scientist, Personal Chef, Fact-Checker, CXO, Web Novelist, Designated Driver, Medical Escort, Hypnotist, Odor Panellist, Career Counselor, Conservator-Restorer, Cartographer, Acoustic Engineer, Insurance Claims Adjuster, Credit Analyst, Headhunter, Optician, Sign Language Interpreter, Pet Trainer, Entomologist, Marine Biologist, Data Annotator, Ergonomist, Real Estate Appraiser, Auctioneer, Library Assistant, Court Reporter, Customer Success Manager, Wedding Planner, Nutrition Coach, Cryptographer, Geophysicist, Technical Writer, Food Scientist, Perfumer, Game Designer, Game Tester, Game Localization Specialist, Business Development Manager, Market Research Analyst, Compensation and Benefits Specialist, Compliance Officer, Chain Store Operations Manager, Merchants Executive, Call Center Agent, Debt Collector, Aerial Photographer, Street Dance Instructor, Luthier/Instrument Maker, Piano Tuner, Stage Designer, Prop Master, Wardrobe Supervisor, Stunt Performer, Script Supervisor, Casting Director, Projectionist, Exhibition Designer, Packaging Designer, Typeface Designer, Brand Strategist, Public Opinion Analyst, Media Buyer, Creative Director, HRBP, Corporate Culture Specialist, Performance Appraisal Specialist, Judicial Auction Specialist, Patent Agent, Trademark Agent, Notary Public, Security Assessor, Emergency Management Specialist, Criminal Psychologist, Handwriting Analyst, Polygraph Examiner, Construction Supervisor, Geotechnical Engineer, Hydraulic Engineer, Port and Waterway Engineer, Tunnel Engineer, Curtain Wall Designer, Blaster, Scaffolder, Fitter, Grinder, Foundry Worker, Forger, Heat Treater, Coating Worker, Forklift Mechanic, Boat Operator, Yacht Captain, Station Master, Parking Attendant, Tire Technician, Tea Master, Food Quality Controller, Hotel Tester, Team-building Coach, Escape Room Designer, Scripted Murder Mystery Host, Relationship Counselor, Pet Detective, Hospice Worker, Genetic Sequencing Analyst, Data Privacy Officer, AI Ethicist.

\subsection{User Personality List}
\subsubsection{User Positive Personality}
\textbf{Social \& Interpersonal.} Warm, Enthusiastic, Cheerful, Outgoing, Friendly, Easygoing, Sincere, Heartfelt, Sincere, Helpful, Empathetic, Understanding, Considerate, Generous, Magnanimous, Generous, Humorous, Witty, Conversational, Talkative, Kind, Affable, Sociable, Charming, Approachable, Modest, Respectful, Polite, Courteous, Loyal, Righteous, Loyal to friends, Reliable, Trustworthy, Cooperative, Team player, Good at socializing, Patient, Tolerant, Forgiving, Inclusive, Tolerant, Kind and honest, Magnanimous, Sunny (personality), Gentle, Hearted, Simple and honest, Unsophisticated, Grateful, Hearty, Frank and open, Frank, Candid, Straightforward.

\textbf{Wisdom \& Mindset.} Smart, Intelligent, Wise, Visionary, Deliberate, Thoughtful, Logical, Methodical, Organized, Analytical, Creative, Imaginative, Curious, Good at learning, Minded, Objective, Rational, Sharp, Astute, Insightful, Resourceful, Flexible, Adaptable, Quick to understand, Wise and farsighted, Witted, Calm, Composed, Headed, Headed, Prudent, Cautious, Erudite, Learned, Inquisitive, Profound thinker, Having one's own opinion, Able to distinguish right from wrong, Pragmatic, Shrewd, Astute, Adaptable, Rigorous, Meticulous, Grained, Attentive to detail.

\textbf{Work Ethic \& Action-Oriented.} Diligent, Hardworking, Dedicated to one's work, Responsible, Conscientious, Serious, Focused, Committed, Invested, Perseverant, Unremitting, Persistent, Disciplined, Efficient, Decisive, Daring, Decisive, Oriented, Proactive, Positive, Proactive, Taking initiative, Enterprising, Ambitious, Ambitious, Constantly striving for perfection, Meticulous, Enduring hardship and hard work, Earth, Oriented, Seeing things through, Punctual, Teachable, Orderly, Methodical, Conscientious and meticulous, To work with abandon, Diligent and conscientious, Steadfast, Dependable, Hardworking, Assiduous, Passionate, Swift and decisive.

\textbf{Moral \& Character.} Honest, Upright, Righteous, Just, Fair, Selfless, Trustworthy, Keeping promises, Incorruptible, Brave, Strong, Firm, Fortitudinous, Humble, Simple, Unadorned, Frugal, Simple, Plain, Hearted, Consistent in thought and action, Open and upright, Noble, Pure, Chaste.

\textbf{Emotional \& Attitude.} Optimistic, Positive and upwardly mobile, Confident, Minded, Magnanimous, Calm and collected, Unhurried, Calm, Resilient, Tenacious, Indomitable, Unyielding, Lively, Energetic, Vibrant, Full of youthful energy, Composed and steady, Mature, Free and easy, Completely at ease, Bold and uninhibited, Independent, Reliant, Mild, Gentle, Quiet, Gentle and quiet, Refined, Cultured, Elegant, Reserved, Implicit, Introverted, Reserved, Healing (personality), Like, Indifferent, Romantic, Interesting, Fun, Having sentimental mood, Appreciates ceremony, Steady, Reliable, Lovely, Cute.

\subsubsection{User Negative Personality}
\textbf{Social \& Interpersonal.} Indifferent, Apathetic, Unsociable, Reclusive, Selfish, Hypocritical, Cunning, Crafty, Treacherous, Deceitful, Mean, Acrimonious, Freeloading, Pinching, Calculating, Rude, Impolite, Arrogant, Boorish, Rude, Domineering, Aggressive, Domineering, Fawning, Sycophantic, Snobbish, Suspicious, Jealous, Vindictive, Holding grudges, Picky, Nitpicky, Stubborn, Tyrannical, Unsociable, Meddlesome, Gossipy, Scheming, Calculating, Inhuman, Inconsiderate, Righteous, Supercilious, Condescending.

\textbf{Wisdom \& Mindset.} Stupid, Foolish, Witted, Dense, Superficial, Dogmatic, Rigid in thinking, Minded, Subjective, Following blindly, Lacking one's own opinion, Muddled, Confused, Obstinate, Fashioned, Track minded, To split hairs, Overly meticulous on trivial matters, To take for granted, Clumsy, Awkward, Ignorant.

\textbf{Work Ethic \& Action-Oriented.} Lazy, Sloppy, Undisciplined, Procrastinating, Perfunctory, Going through the motions, Careless, Careless, Irresponsible, Giving up halfway, Opportunistic, Aiming too high, Unrealistic, High standards but low ability, Disorganized, Sticking to old conventions, Idle, Loafing, Lived enthusiasm.

\textbf{Moral \& Character.} Vain, Greedy, Seeking, Cowardly, Timid, Extravagant, Wasteful, Loving leisure and hating labor.

\textbf{Emotional \& Attitude.} Pessimistic, Negative, Anxious, Depressed, Gloomy, Tempered, Cranky, Irascible, Emotional, Moody, Overly sensitive, Fragile, Glass heart, Easily offended, Melancholy, Neurotic, Paranoid, Complaining and blaming others, Esteem, Impulsive, Impetuous, Restless, Sentimental, Melancholic, Moody, Unpredictable.


\section{Experiment Setting}
\label{experiment_setting}
In this section, we detail the specific training configurations. We employ the SWIFT framework for both the Supervised Fine-Tuning (SFT) and Reinforcement Learning (RL) stages. All experiments are conducted on 16 NVIDIA H100 GPUs, utilizing Qwen3VL-8B-Instruct as the backbone model. 
\subsection{SFT Setting}
For the SFT stage, we perform full-parameter fine-tuning on the language model backbone, while keeping the vision encoder frozen. The learning rate is initialized at $1e^{-6}$ (decaying to a minimum of $1e^{-7}$), with a batch size of 16 and a maximum context length of 128K tokens. We train for 2 epochs, limiting the maximum number of pixels to 1,003,520 pixels. For distributed training, we configure a tensor parallelism (TP) size of 8 and a pipeline parallelism (PP) size of 2.
\subsection{RL Setting}
In the RL stage, we employ LoRA for parameter-efficient fine-tuning, configured with a rank $r=32$ and a scaling factor $\alpha=128$. We initialize the learning rate at $1e^{-6}$, with a batch size of 64 and a maximum context length of 20K tokens. For sequence-level optimization, we adopt the GSPO algorithm, setting the group size to 16, the KL regularization coefficient to 0.001, and the clipping range to $\pm 0.2$. The model is trained for 1 epoch with the image resolution capped at 200,704 pixels. Finally, we utilize a distributed setup with a tensor parallelism size of 8 and a pipeline parallelism size of 1.


\section{Objective Metrics Definitions And Computations}
\label{objective_metrics_computation}
\subsection{Definitions And Computations}
In our evaluation framework, we distinguish between \textit{Event-driven} and \textit{Intent-driven} triggering mechanisms. Consequently, the definitions and computations of True Positive (TP), False Positive (FP), True Negative (TN), and False Negative (FN) differ to accommodate the granularities of frame-level streams versus segment-level intentions.

\subsubsection{The Definitions And Computations Of TP, FP, TN, FN In Event-driven Evaluations.}
For streaming event triggering, the evaluation is performed continuously at the frame or time-step level. The model outputs a binary response (\textit{\textless Attention\textgreater} or \textit{\textless Silence\textgreater}) for every input frame.

\textbf{TP.} A sample is considered a True Positive when the model correctly predicts the occurrence of an event (output \textit{\textless Attention\textgreater}) at a time step where the ground truth also indicates an event (label \textit{\textless Attention\textgreater}).

\textbf{FP.} A False Positive occurs when the model predicts an event (output \textit{\textless Attention\textgreater}) at a time step where the ground truth indicates no event (label \textit{\textless Silence\textgreater}). This represents a false alert.

\textbf{TN.} A True Negative is recorded when the model correctly predicts the absence of an event (output \textit{\textless Silence\textgreater}) at a time step where the ground truth also indicates no event (label \textit{\textless Silence\textgreater})

\textbf{FN.} A False Negative occurs when the model fails to predict an existing event (output \textit{\textless Silence\textgreater}) at a time step where the ground truth indicates the presence of an event (label \textit{\textless Attention\textgreater})

\subsubsection{The Definitions And Computations Of TP, FP, TN, FN In Intent-driven Evaluations.}
For streaming intent triggering, the model generates discrete response segments that must be aligned with ground truth (GT) intervals. We employ the Hungarian matching algorithm to optimally assign predicted segments to GT segments based on temporal overlap.

\textbf{TP.} A predicted intent segment is classified as a True Positive if it is successfully matched to a ground truth interval via the Hungarian algorithm. This implies the prediction correctly identifies an intent within a valid temporal proximity to the GT.

\textbf{FP.} A False Positive is defined as a predicted intent segment that fails to match any ground truth interval after the Hungarian allocation. These are considered surplus or incorrect predictions.

\textbf{TN.} In the context of intent segments, True Negatives represent the correctly identified absence of intents. While traditionally elusive in detection tasks, this contributes to the overall accuracy calculation by accounting for non-intent periods where no predictions were triggered.

\textbf{FN.} A False Negative corresponds to a ground truth intent interval that remains unmatched after the assignment process. This signifies that the model failed to trigger a response for a genuine user intent.

\subsection{Metrics Computation}

Precision = $\frac{TP}{TP + FP}$

Recall = $\frac{TP}{TP + FN}$

F1-score = $2 \times \frac{Precision \times Recall}{Precision + Recall}$

\textbf{Ground-truth Hit Accuracy.} This metric evaluates the system's coverage capability. It is calculated as the ratio of ground truth segments that are successfully "hit" (matched) by at least one prediction to the total number of ground truth segments. This ensures that we measure not just how precise the predictions are, but how exhaustively the model retrieves all distinct events or intents.

\textbf{mIoU.} We compute the temporal Intersection over Union (IoU) to evaluate the alignment between predicted response durations and ground truth intervals. For \textit{Event-driven triggering}, it is computed as the standard Intersection over Union (IoU) score between the predicted binary event stream and the ground truth event stream. Specifically, it calculates the ratio of the temporal intersection of the 'alert' intervals to their temporal union. As for \textit{Intent-driven triggering}, the calculation follows the same principle. After aligning the predicted intent segments with the ground truth intervals (using the aforementioned Hungarian matching strategy for assignment), we calculate the IoU to quantify the precise temporal overlap between the response and the user's actual intent duration.

\textbf{LLM-as-a-Judge Score Computation Strategy.} To derive the final values for these quality metrics, we implement a rigorous aggregation strategy that isolates generation quality from retrieval performance. Scores are computed exclusively for instances where both the ground truth (GT) and the model prediction successfully trigger an ``alert" state. In scenarios where multiple prediction segments correspond to a single ground truth interval, we adopt a conservative approach by assigning the \textit{lowest} score among them to that interval, thereby penalizing inconsistent multiple responses. The final system-level score is calculated as the mean of these representative scores across all matched GT intervals, explicitly excluding ground truth intervals that fail to elicit a response (false negatives) to ensure the metric purely reflects the quality of generated content independent of the system's recall capabilities.

\textbf{Memory Consistency.} To assess the quality of model responses in intent understanding tasks, we introduce Memory Consistency as a key LLM-based evaluation metric. This metric measures the alignment between the model's response and the user's long-term memory, such as established preferences and background context. The evaluation adheres to a streaming frame-by-frame protocol utilizing a local context window. We utilize a 0-to-5 scoring scale. 


\textbf{Response Quality.} Response Quality focuses on the effectiveness of the content, specifically evaluating whether the response provides actionable alerts or practical suggestions. Similar to Memory Consistency, this metric employs an LLM-based judge using a 0-to-5 scoring scale under the streaming protocol.


\section{Full Results Of EgoPro-Intent}
\label{intent_driven_detals_result}
%

\small

\begin{longtable}{l l *{7}{c}}
\caption{Detailed results for each domain on the EgoPro-Intent subset.}
\label{tab:intent_driven_detals_result} \\

\toprule
\textbf{Model} & \textbf{Domain} & \multicolumn{7}{c}{EgoPro-Intent} \\
\cmidrule(lr){3-9}
 &  & Precision & Recall & F1. & GHA & mIoU & MC & RQ \\
\midrule
\endfirsthead

\toprule
\textbf{Model} & domain & \multicolumn{7}{c}{Intent} \\
\cmidrule(lr){3-9}
 &  & Precision & Recall & F1. & GHA & mIoU & MC & RQ \\
\midrule
\endhead

\midrule
\multicolumn{9}{r}{Continued on next page} \\
\midrule
\endfoot

\bottomrule
\endlastfoot

\rowcolor{red!5}
\multicolumn{9}{l}{\textbf{VideoChat-R1.5}} \\
 & Cooking & 7.39 & 78.50 & 12.46 & 68.85 & 60.92 & 3.39 & 1.55 \\
 & Dailylife & 8.88 & 64.09 & 14.69 & 59.76 & 47.06 & 2.54 & 1.21 \\
 & Driving & 8.45 & 65.03 & 14.07 & 54.79 & 43.33 & 2.88 & 1.42 \\
 & Entertainment & 6.63 & 74.45 & 11.41 & 59.12 & 57.00 & 3.62 & 1.49 \\
 & Navigation & 8.10 & 71.67 & 13.07 & 60.35 & 32.09 & 3.21 & 1.05 \\
 & Art & 7.10 & 78.14 & 11.77 & 58.98 & 61.37 & 3.33 & 1.95 \\
 & Shopping & 9.63 & 86.29 & 16.56 & 68.26 & 58.73 & 3.82 & 1.93 \\
 & Sports & 8.01 & 76.79 & 13.54 & 56.75 & 61.82 & 3.64 & 1.76 \\
 & Travel & 10.74 & 88.91 & 18.30 & 58.88 & 65.08 & 3.59 & 2.18 \\
 & Working & 6.92 & 77.11 & 12.20 & 59.63 & 57.12 & 3.34 & 1.62 \\
\rowcolor{blue!5}  & Mean & 8.19 & 76.09 & 13.82 & 60.54 & 54.44 & 3.34 & 1.62 \\
\hline
\rowcolor{red!5}
\multicolumn{9}{l}{\textbf{VideoRFT-7B}} \\
 & Cooking & 7.05 & 94.75 & 12.71 & 55.49 & 72.14 & 4.31 & 1.57 \\
 & Dailylife & 10.60 & 93.08 & 18.19 & 55.65 & 67.75 & 3.79 & 1.54 \\
 & Driving & 11.05 & 99.50 & 19.13 & 55.92 & 64.58 & 4.05 & 1.94 \\
 & Entertainment & 6.74 & 94.05 & 12.25 & 54.46 & 71.40 & 4.26 & 1.86 \\
 & Navigation & 7.63 & 100.00 & 13.77 & 46.79 & 40.89 & 4.56 & 1.31 \\
 & Art & 6.75 & 90.62 & 12.19 & 56.91 & 72.57 & 3.84 & 2.30 \\
 & Shopping & 9.25 & 100.00 & 16.39 & 50.44 & 62.02 & 4.67 & 1.90 \\
 & Sports & 8.99 & 90.50 & 15.83 & 55.59 & 69.37 & 4.18 & 2.25 \\
 & Travel & 10.53 & 99.43 & 18.52 & 57.45 & 72.97 & 4.25 & 2.52 \\
 & Working & 7.80 & 88.87 & 13.84 & 53.37 & 62.78 & 3.99 & 1.94 \\
\rowcolor{blue!5}  & Mean & 8.64 & 95.11 & 15.29 & 54.21 & 65.66 & 4.19 & 1.91 \\
\hline
\rowcolor{red!5}
\multicolumn{9}{l}{\textbf{Qwen2.5vl-3B}} \\
 & Cooking & 7.15 & 100.00 & 12.96 & 49.47 & 75.06 & 3.88 & 1.58 \\
 & Dailylife & 11.29 & 100.00 & 19.42 & 51.93 & 71.64 & 3.28 & 1.51 \\
 & Driving & 11.11 & 100.00 & 19.24 & 55.00 & 64.79 & 3.39 & 1.43 \\
 & Entertainment & 7.15 & 100.00 & 13.00 & 53.73 & 75.35 & 4.27 & 1.63 \\
 & Navigation & 7.63 & 100.00 & 13.76 & 46.62 & 40.88 & 4.04 & 1.22 \\
 & Art & 7.16 & 100.00 & 12.95 & 58.60 & 79.92 & 3.97 & 2.09 \\
 & Shopping & 9.13 & 100.00 & 16.21 & 47.87 & 61.78 & 4.39 & 1.83 \\
 & Sports & 9.40 & 100.00 & 16.71 & 55.03 & 75.01 & 4.16 & 1.83 \\
 & Travel & 10.54 & 100.00 & 18.54 & 57.25 & 73.51 & 3.69 & 1.86 \\
 & Working & 8.49 & 100.00 & 15.16 & 51.43 & 69.21 & 3.96 & 1.83 \\
\rowcolor{blue!5}  & Mean & 8.91 & 100.00 & 15.80 & 52.70 & 68.71 & 3.90 & 1.68 \\
\hline
\rowcolor{red!5}
\multicolumn{9}{l}{\textbf{Qwen2.5VL-7B}} \\
 & Cooking & 6.59 & 89.38 & 11.88 & 54.36 & 67.42 & 3.78 & 1.68 \\
 & Dailylife & 10.78 & 91.52 & 18.20 & 55.39 & 66.77 & 3.33 & 1.82 \\
 & Driving & 11.11 & 100.00 & 19.24 & 55.00 & 64.79 & 3.99 & 2.17 \\
 & Entertainment & 7.17 & 98.50 & 12.99 & 55.17 & 74.23 & 4.41 & 2.08 \\
 & Navigation & 7.63 & 99.33 & 13.75 & 46.96 & 40.53 & 4.21 & 1.50 \\
 & Art & 7.11 & 97.50 & 12.85 & 59.71 & 77.92 & 3.90 & 2.48 \\
 & Shopping & 9.20 & 98.50 & 16.22 & 48.78 & 60.81 & 4.42 & 2.44 \\
 & Sports & 9.26 & 98.08 & 16.46 & 55.45 & 73.95 & 4.31 & 2.31 \\
 & Travel & 10.54 & 100.00 & 18.53 & 57.25 & 73.51 & 4.04 & 2.39 \\
 & Working & 8.25 & 97.16 & 14.69 & 52.75 & 67.61 & 4.23 & 2.19 \\
\rowcolor{blue!5}  & Mean & 8.76 & 97.00 & 15.48 & 54.09 & 66.75 & 4.06 & 2.10 \\
\hline
\rowcolor{red!5}
\multicolumn{9}{l}{\textbf{Qwen2.5VL-32B}} \\
 & Cooking & 7.21 & 99.75 & 13.04 & 54.21 & 75.44 & 3.88 & 1.77 \\
 & Dailylife & 11.51 & 99.50 & 19.72 & 58.45 & 71.92 & 3.83 & 2.09 \\
 & Driving & 11.13 & 98.98 & 19.23 & 56.42 & 64.40 & 4.19 & 2.38 \\
 & Entertainment & 7.26 & 99.83 & 13.17 & 57.79 & 75.62 & 4.41 & 2.37 \\
 & Navigation & 7.79 & 98.47 & 14.01 & 47.92 & 40.73 & 4.25 & 1.37 \\
 & Art & 7.16 & 99.58 & 12.96 & 61.43 & 80.22 & 3.98 & 2.41 \\
 & Shopping & 9.35 & 99.94 & 16.54 & 55.25 & 62.57 & 4.12 & 2.42 \\
 & Sports & 9.52 & 100.00 & 16.90 & 56.92 & 74.60 & 4.40 & 2.28 \\
 & Travel & 10.30 & 100.00 & 18.16 & 57.23 & 73.30 & 4.07 & 2.37 \\
 & Working & 8.65 & 99.44 & 15.33 & 53.20 & 66.92 & 4.27 & 2.50 \\
\rowcolor{blue!5}  & Mean & 8.95 & 99.54 & 15.84 & 55.96 & 68.49 & 4.13 & 2.18 \\
\hline
\rowcolor{red!5}
\multicolumn{9}{l}{\textbf{Qwen2.5VL-72B}} \\
 & Cooking & 7.28 & 98.40 & 13.09 & 55.00 & 74.77 & 3.83 & 1.91 \\
 & Dailylife & 11.51 & 95.08 & 19.42 & 59.94 & 70.07 & 3.15 & 1.67 \\
 & Driving & 12.90 & 98.75 & 20.55 & 58.52 & 65.73 & 4.21 & 1.93 \\
 & Entertainment & 7.21 & 99.33 & 13.09 & 58.27 & 75.66 & 4.15 & 2.11 \\
 & Navigation & 7.63 & 99.39 & 13.76 & 47.55 & 40.71 & 4.20 & 1.51 \\
 & Art & 7.21 & 99.75 & 13.03 & 61.37 & 80.35 & 3.96 & 2.09 \\
 & Shopping & 10.20 & 99.30 & 17.41 & 58.67 & 63.65 & 4.12 & 2.18 \\
 & Sports & 9.70 & 99.58 & 17.00 & 59.59 & 75.42 & 4.28 & 2.11 \\
 & Travel & 10.64 & 99.95 & 18.69 & 59.22 & 73.80 & 3.69 & 2.22 \\
 & Working & 9.08 & 99.02 & 15.35 & 55.71 & 68.73 & 3.97 & 2.06 \\
\rowcolor{blue!5}  & Mean & 9.34 & 98.86 & 16.14 & 57.39 & 68.89 & 3.95 & 1.98 \\
\hline
\rowcolor{red!5}
\multicolumn{9}{l}{\textbf{Qwen3VL-4B}} \\
 & Cooking & 11.85 & 58.04 & 14.72 & 68.02 & 47.81 & 2.89 & 1.78 \\
 & Dailylife & 16.61 & 64.55 & 20.76 & 70.09 & 51.34 & 2.70 & 1.85 \\
 & Driving & 13.43 & 44.23 & 15.92 & 59.46 & 34.81 & 2.19 & 1.26 \\
 & Entertainment & 11.68 & 72.62 & 15.91 & 72.58 & 60.60 & 3.68 & 2.16 \\
 & Navigation & 8.21 & 87.73 & 14.24 & 61.21 & 38.94 & 4.04 & 1.32 \\
 & Art & 9.60 & 93.86 & 15.05 & 69.05 & 77.54 & 4.20 & 2.95 \\
 & Shopping & 17.88 & 74.94 & 23.79 & 73.64 & 53.66 & 3.80 & 2.50 \\
 & Sports & 15.82 & 81.54 & 21.50 & 72.11 & 68.17 & 4.17 & 2.41 \\
 & Travel & 18.75 & 81.35 & 24.70 & 70.04 & 64.64 & 3.87 & 2.68 \\
 & Working & 9.11 & 72.06 & 13.84 & 67.07 & 54.43 & 3.35 & 2.13 \\
\rowcolor{blue!5}  & Mean & 13.29 & 73.09 & 18.04 & 68.33 & 55.19 & 3.49 & 2.10 \\
\hline
\rowcolor{red!5}
\multicolumn{9}{l}{\textbf{Qwen3VL-30B-A3B}} \\
 & Cooking & 8.45 & 96.08 & 14.52 & 73.80 & 76.89 & 4.36 & 2.71 \\
 & Dailylife & 13.59 & 90.74 & 21.39 & 68.64 & 68.54 & 3.72 & 2.54 \\
 & Driving & 11.69 & 78.44 & 19.02 & 62.36 & 55.14 & 4.01 & 2.61 \\
 & Entertainment & 9.55 & 91.42 & 15.46 & 66.81 & 72.37 & 4.56 & 2.81 \\
 & Navigation & 8.29 & 84.79 & 14.30 & 64.46 & 36.66 & 3.79 & 1.33 \\
 & Art & 7.37 & 95.79 & 13.22 & 64.75 & 77.19 & 4.25 & 3.23 \\
 & Shopping & 11.50 & 93.75 & 18.85 & 72.36 & 62.68 & 4.20 & 2.73 \\
 & Sports & 10.01 & 89.02 & 16.79 & 63.25 & 72.47 & 4.71 & 2.88 \\
 & Travel & 12.04 & 94.17 & 20.14 & 64.83 & 69.59 & 4.33 & 3.20 \\
 & Working & 8.73 & 89.98 & 15.05 & 63.94 & 67.39 & 4.22 & 2.67 \\
\rowcolor{blue!5}  & Mean & 10.13 & 90.42 & 16.88 & 66.53 & 65.88 & 4.21 & 2.67 \\
\hline
\rowcolor{red!5}
\multicolumn{9}{l}{\textbf{Qwen3VL-8B}} \\
 & Cooking & 15.95 & 83.94 & 20.62 & 75.46 & 66.20 & 4.08 & 2.59 \\
 & Dailylife & 19.58 & 75.66 & 23.59 & 69.62 & 56.12 & 3.34 & 2.10 \\
 & Driving & 14.39 & 55.47 & 17.86 & 59.77 & 40.17 & 3.44 & 2.02 \\
 & Entertainment & 14.70 & 84.62 & 18.35 & 72.85 & 71.15 & 4.48 & 2.88 \\
 & Navigation & 6.96 & 67.97 & 11.81 & 60.27 & 29.54 & 3.09 & 1.15 \\
 & Art & 16.07 & 85.82 & 20.45 & 70.74 & 71.14 & 4.12 & 2.72 \\
 & Shopping & 14.25 & 79.65 & 20.07 & 72.85 & 54.76 & 4.02 & 2.48 \\
 & Sports & 18.91 & 74.21 & 22.90 & 70.26 & 62.00 & 4.24 & 2.58 \\
 & Travel & 19.24 & 78.88 & 25.99 & 67.61 & 60.46 & 3.91 & 2.77 \\
 & Working & 12.86 & 79.26 & 17.73 & 67.35 & 61.30 & 4.02 & 2.24 \\
\rowcolor{blue!5}  & Mean & 15.30 & 76.50 & 19.94 & 68.68 & 57.21 & 3.87 & 2.35 \\
\hline
\rowcolor{red!5}
\multicolumn{9}{l}{\textbf{TimeChatOnline-7B}} \\
 & Cooking & 5.06 & 64.54 & 8.51 & 53.52 & 49.08 & 2.83 & 1.20 \\
 & Dailylife & 7.76 & 60.04 & 12.62 & 53.87 & 44.28 & 2.36 & 1.08 \\
 & Driving & 10.96 & 98.25 & 18.95 & 55.52 & 63.72 & 3.89 & 1.77 \\
 & Entertainment & 5.63 & 75.67 & 10.18 & 56.22 & 58.28 & 3.46 & 1.33 \\
 & Navigation & 7.50 & 98.50 & 13.54 & 46.60 & 40.37 & 4.30 & 1.47 \\
 & Art & 6.24 & 85.83 & 11.34 & 57.46 & 68.64 & 3.65 & 2.12 \\
 & Shopping & 6.16 & 63.94 & 10.89 & 52.22 & 40.57 & 2.87 & 1.40 \\
 & Sports & 8.73 & 89.25 & 15.48 & 55.13 & 66.42 & 4.13 & 1.95 \\
 & Travel & 10.06 & 95.93 & 17.72 & 56.73 & 70.46 & 3.85 & 2.07 \\
 & Working & 6.70 & 73.66 & 11.87 & 53.22 & 52.08 & 3.22 & 1.57 \\
\rowcolor{blue!5}  & Mean & 7.48 & 80.59 & 13.11 & 54.05 & 55.40 & 3.46 & 1.60 \\
\hline
\rowcolor{red!5}
\multicolumn{9}{l}{\textbf{SFT (Ours)}} \\
 & Cooking & 52.49 & 41.35 & 42.95 & 66.35 & 62.16 & 3.12 & 2.38 \\
 & Dailylife & 78.10 & 58.15 & 62.38 & 75.54 & 63.85 & 3.80 & 3.05 \\
 & Driving & 81.43 & 52.74 & 60.88 & 73.50 & 56.85 & 4.17 & 3.27 \\
 & Entertainment & 81.37 & 58.54 & 64.25 & 76.76 & 60.85 & 4.41 & 3.62 \\
 & Navigation & 35.90 & 26.74 & 28.78 & 65.39 & 32.96 & 1.90 & 1.19 \\
 & Art & 89.82 & 63.17 & 69.11 & 77.90 & 61.24 & 4.73 & 4.01 \\
 & Shopping & 53.69 & 31.31 & 35.42 & 62.74 & 45.19 & 2.92 & 2.31 \\
 & Sports & 84.64 & 56.33 & 64.23 & 75.85 & 56.02 & 4.53 & 3.57 \\
 & Travel & 85.47 & 67.15 & 71.23 & 80.00 & 71.85 & 4.48 & 3.76 \\
 & Working & 74.63 & 45.25 & 50.99 & 69.36 & 51.85 & 4.01 & 3.06 \\
\rowcolor{blue!5}  & Mean & 71.75 & 50.07 & 55.02 & 72.34 & 56.28 & 3.81 & 3.02 \\
\hline
\rowcolor{red!5}
\multicolumn{9}{l}{\textbf{RL W/O Think (Ours)}} \\
 & Cooking & 54.99 & 43.01 & 44.90 & 67.50 & 62.12 & 3.20 & 2.43 \\
 & Dailylife & 78.40 & 57.21 & 61.99 & 74.99 & 63.50 & 3.78 & 3.02 \\
 & Driving & 81.43 & 52.49 & 60.71 & 73.34 & 56.05 & 4.17 & 3.21 \\
 & Entertainment & 80.82 & 58.86 & 64.21 & 76.96 & 61.19 & 4.37 & 3.63 \\
 & Navigation & 34.82 & 25.82 & 27.84 & 64.86 & 32.08 & 1.77 & 1.11 \\
 & Art & 89.57 & 62.75 & 68.63 & 77.67 & 60.86 & 4.74 & 3.99 \\
 & Shopping & 54.44 & 32.43 & 36.12 & 63.44 & 45.34 & 2.96 & 2.32 \\
 & Sports & 85.35 & 57.41 & 65.18 & 76.44 & 56.11 & 4.50 & 3.55 \\
 & Travel & 86.45 & 66.91 & 71.55 & 79.85 & 71.82 & 4.54 & 3.77 \\
 & Working & 75.25 & 44.23 & 50.25 & 68.85 & 50.58 & 4.08 & 3.06 \\
\rowcolor{blue!5}  & Mean & 72.15 & 50.11 & 55.14 & 72.39 & 55.96 & 3.81 & 3.01 \\
\hline
\rowcolor{red!5}
\multicolumn{9}{l}{\textbf{RL W Think (Ours)}} \\
 & Cooking & 52.69 & 59.14 & 51.27 & 75.19 & 71.08 & 3.92 & 3.11 \\
 & Dailylife & 70.66 & 59.92 & 60.28 & 76.97 & 62.27 & 3.90 & 2.92 \\
 & Driving & 74.15 & 64.17 & 63.10 & 79.20 & 60.76 & 4.62 & 3.76 \\
 & Entertainment & 67.22 & 65.69 & 60.59 & 79.75 & 68.36 & 4.53 & 3.62 \\
 & Navigation & 36.83 & 32.20 & 31.05 & 66.50 & 46.35 & 2.20 & 1.11 \\
 & Art & 85.12 & 65.60 & 67.26 & 79.63 & 64.31 & 4.88 & 4.23 \\
 & Shopping & 53.56 & 50.28 & 46.64 & 72.50 & 56.45 & 3.66 & 2.78 \\
 & Sports & 74.86 & 64.49 & 63.87 & 79.86 & 64.16 & 4.59 & 3.79 \\
 & Travel & 82.30 & 70.23 & 71.22 & 81.51 & 70.63 & 4.66 & 3.96 \\
 & Working & 67.61 & 46.89 & 48.06 & 70.14 & 51.83 & 4.02 & 3.07 \\
\rowcolor{blue!5}  & Mean & 66.50 & 57.86 & 56.34 & 76.13 & 61.62 & 4.10 & 3.23 \\

\end{longtable}




\end{document}